\documentclass[conference]{IEEEtran}
\usepackage{times}

\usepackage[numbers]{natbib}
\usepackage{multicol}
\usepackage[bookmarks=true]{hyperref}
\usepackage{amsmath}
\usepackage{subcaption}
\usepackage{hyperref}
\usepackage{amsmath} %
\usepackage{amsfonts}
\usepackage{mathtools}
\usepackage{cleveref}
\hypersetup{
    colorlinks=true,
    linkcolor=black,
    citecolor = black,
    filecolor=magenta,      
    urlcolor=blue
    }
\usepackage{aux_files/tikz_styles}

\usepackage{changes}
\definechangesauthor[name={S.~M.}, color={blue}]{sm}
\definechangesauthor[name={T.~M.}, color={orange}]{tm}
\definechangesauthor[name={A.~A.}, color={purple}]{aa}
\definechangesauthor[name={B.P.~G.}, color={brown}]{bpg}
\definechangesauthor[name={R.~T.}, color={red}]{rt}
\definechangesauthor[name={P.~P.}, color={magenta}]{pp}

\newtheorem{example}{Example}

\begin{document}
\title{Mixed Discrete and Continuous Planning using \\ Shortest Walks in Graphs of Convex Sets}

\author{
  \authorblockN{Savva Morozov\textsuperscript{1},
    Tobia Marcucci\textsuperscript{2},
    Bernhard Paus Graesdal\textsuperscript{1},
    Alexandre Amice\textsuperscript{1},
    Pablo Parrilo\textsuperscript{1},
    Russ Tedrake\textsuperscript{1,3}} 
  \authorblockA{\textsuperscript{1}Massachusetts Institute of Technology}
  \authorblockA{\textsuperscript{2}University of California, Santa Barbara}
  \authorblockA{\textsuperscript{3}Toyota Research Institute\\
  \texttt{savva@mit.edu, marcucci@ucsb.edu, \{graesdal, amice, parrilo, russt\}@mit.edu}
  }
}

\maketitle

\begin{abstract}
We study the Shortest-Walk Problem (SWP) in a Graph of Convex Sets (GCS).
A GCS is a graph where each vertex is paired with a convex program, and each edge couples adjacent programs via additional costs and constraints.
A walk in a GCS is a sequence of vertices connected by edges, where vertices may be repeated.
The length of a walk is given by the cumulative optimal value of the corresponding convex programs.
To solve the SWP in GCS, we first synthesize a piecewise-quadratic lower bound on the problem's cost-to-go function using semidefinite programming.
Then we use this lower bound to guide an incremental-search algorithm that yields an approximate shortest walk.
We show that the SWP in GCS is a natural language for many mixed discrete-continuous planning problems in robotics, unifying problems that typically require specialized solutions while delivering high performance and computational efficiency. 
We demonstrate this through experiments in collision-free motion planning, skill chaining, and optimal control of hybrid systems.

\end{abstract}

\IEEEpeerreviewmaketitle

\section{Introduction}

A Graph of Convex Sets (GCS) is a generalization of a directed graph where each vertex is paired with a convex convex program, and each edge couples adjacent programs with additional costs and constraints~\cite{marcucci2024graphs}.
When traversing a GCS, we must select a feasible point for the program of each vertex that we visit, and these points must also verify the constraints paired with the traversed edge.
The total cost of the traversal is the sum of the costs of these vertices and edges.
Many classical problems in graph theory, such as the shortest-path problem, the traveling-salesman problem, the minimum-spanning-tree problem, are naturally extended to a GCS.
Among these, the Shortest-Path Problem (SPP) in GCS~\cite{marcucci2024shortest} has received particular attention due to its many applications in robotics. 
The objective is to find a discrete path through the graph, together with the continuous vertex points along this path, that minimize the cumulative cost.
The SPP in GCS naturally models problems where discrete and continuous decision-making are interleaved, making it a powerful tool for robotics applications, including optimal control~\cite{marcucci2024shortest}, collision-free motion planning~\cite{marcucci2023motion,cohn2023non,von2024using}, planning through contact~\cite{graesdal2024towards}, and other problems~\cite{philip2024mixed,kurtz2023temporal}.

\begin{figure}[t!]
    \centering
    \includegraphics[width=1.0\columnwidth]{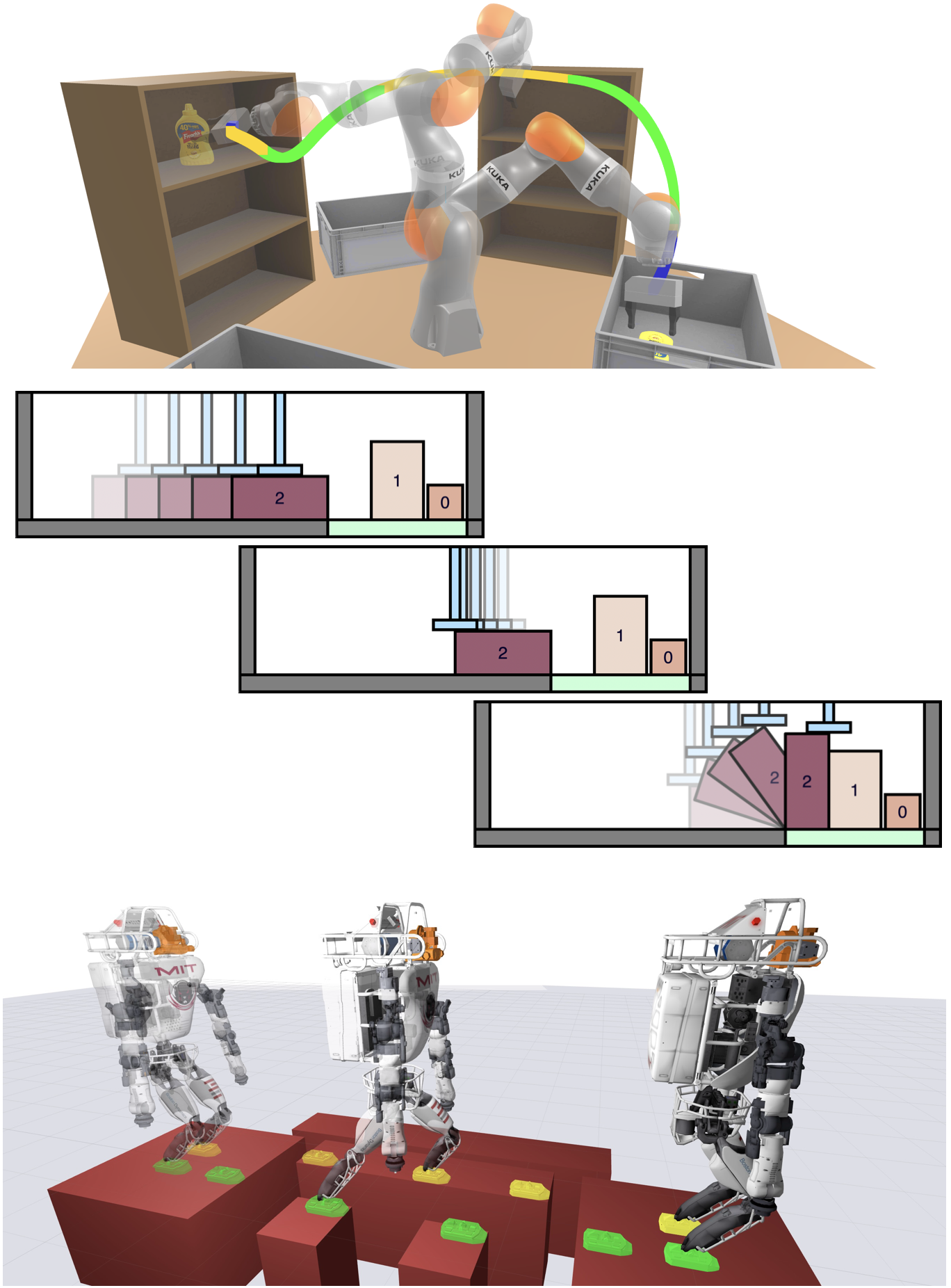}
    \caption{
Application of the SWP in GCS to a variety of robotics planning problems: collision-free motion planning for a robot arm with derivative constraints, item sorting for a top-down suction gripper, and footstep planning for a humanoid robot.
    }
    \label{f-front-image}
\end{figure}

In this paper, we study the Shortest-Walk Problem (SWP) in GCS: instead of searching for a path through the graph, which is a sequence of \emph{distinct} vertices, we search for a walk, which allows vertex revisits. 
It is well-known that for an ordinary graph with non-negative costs, the SWP reduces to the SPP, since revisiting a vertex only adds a cycle with non-negative cost, and makes no progress towards the target.
Conversely, the SWP and the SPP are significantly different for a GCS.
By allowing vertex revisits, the SWP captures a wide range of problems in robotics and control more naturally, including collision-free motion planning with derivative constraints, skill chaining, and optimal control of hybrid dynamical systems.
Our experimental results highlight that this problem is widely applicable in robotic manipulation and locomotion, as illustrated in \Cref{f-front-image}.

To solve the SWP in GCS, we extend the methodology proposed in~\cite{morozov2024multi} for the SPP in GCS.
First, we synthesize a piecewise-quadratic lower bound on the problem's cost-to-go function, with each quadratic piece defined over the convex set associated with each GCS vertex.
Next, we use this lower bound to guide a search algorithm, obtaining a walk incrementally, one vertex at a time.
Though we can produce provably optimal walks, in practice we use a faster heuristic method that quickly yields effective solutions.

This paper is organized as follows.
We formulate the SWP in GCS in \Cref{s-problem-statement}, highlighting its differences with respect to the SPP in GCS.
In \Cref{s-many-problems-cast}, we motivate our study of this problem by demonstrating its relevance for real-world applications in robotics and optimal control.
We derive a practical numerical approach for solving this problem in \Cref{s-solution-method}.
We demonstrate the performance of our approach across a variety of experimental domains in \Cref{s-results} and discuss its limitations in \Cref{s-limitations}.

\section{Shortest Walks in Graphs of Convex Sets}
\label{s-problem-statement}
In this section, we define the SWP in GCS and highlight the main differences with respect to the SPP in GCS. 
For more details  and other GCS problems, we refer the reader to~\cite{marcucci2024graphs}.

\subsection{Preliminaries}
\paragraph*{Graph of Convex Sets (GCS)}
A GCS is a directed graph $G = (\mathcal V, \mathcal E)$ with vertex set $\mathcal V$ and edge set $\mathcal E$.
Each vertex $v\in\mathcal V$ is paired with a convex program, defined by a compact convex set $\mathcal X_v$ and a non-negative convex cost function ${l_v: \mathcal{X}_v \rightarrow \mathbb{R}_+}$.
Similarly, each edge $e = (u,v)\in\mathcal E$ is paired with a convex set ${\mathcal X_e \subseteq \mathcal{X}_u \times \mathcal{X}_v}$ and a non-negative convex cost function ${l_e: \mathcal{X}_e \rightarrow \mathbb{R}_+}$. 
When traversing a GCS, we must select a point $x_v\in\mathcal X_v$ upon a visit to vertex $v$ and incur the cost $l_v(x_v)$.
When moving along an edge $e=(u,v)$, the adjacent points $(x_u,x_v)$ must satisfy the constraint $(x_u,x_v)\in\mathcal X_e$, and we incur the edge cost $l_e(x_u,x_v)$.

\paragraph*{Walk in a GCS} Let $s,t$ be a pair of source and target vertices, and $\bar x_s\in\mathcal X_s$, $\bar x_t\in\mathcal X_t$ be a pair of source and target points.
A \textit{$K$-step, $s\textsf{-}t$ walk in a GCS} between points $\bar x_s$ and $\bar x_t$ is a sequence of $K+1$ vertices ${w=(v_0,\ldots, v_K)}$ and a sequence of $K+1$ vertex points ${\tau=(x_0,\ldots, x_K)}$ such that
\begin{subequations}\label{e-walk-def}
\begin{align}
& v_0=s, \; v_K=t,\label{e-walk-vb}\\
& x_0=\bar x_s, \; x_K = \bar x_t,\label{e-walk-pb}\\
& e_k = (v_{k-1}, v_{k})\in\mathcal E,   \quad & \forall k=1,\ldots, K, \label{e-walk-vc}\\
& (x_{k-1},x_{k}) \in \mathcal X_{e_k}, & \forall k=1,\ldots, K. \label{e-walk-pc}
\end{align}
\end{subequations}
In words, we require that the walk start at vertex $s$ and point $\bar x_s$, and end at vertex $t$ and point $\bar x_t$.
Consecutive pairs of vertices $(v_{k-1}, v_k)$ must be connected by an edge $e_k\in\mathcal E$, and consecutive pairs of points $(x_{k-1},x_k)$ must lie in the corresponding edge constraint set $\mathcal X_{e_k}$.
The later also ensures that the vertex constraints are satisfied along the walk, since ${\mathcal X_{e_k} \subseteq \mathcal{X}_{v_{k-1}} \times \mathcal{X}_{v_k}}$ by definition.
The tuple $(w,\tau)$ is the walk in the GCS.
Individually, the sequence of vertices $w$ is a walk in the graph $G$, while $\tau$ is the corresponding sequence of vertex points, referred to as the \textit{trajectory}.
We denote the set of $K$-step walks $w$ that satisfy \eqref{e-walk-vb}, \eqref{e-walk-vc} as $\mathcal W^K_{s,t}$.
For a given $w$, we denote the set of trajectories $\tau$ that satisfy \eqref{e-walk-pb}, \eqref{e-walk-pc} as $\mathcal T_w(\bar x_s, \bar x_t)$.
We emphasize that vertices and edges along the walk $w$ may be repeated, and that different continuous points may be selected upon revisiting the same vertex.

We denote $l(w,\tau)$ as the sum of edge and vertex costs along the walk $(w,\tau)$, referring to it as the cost of a walk in a GCS:
$$
l(w,\tau) = \sum_{k=0}^{K}l_{v_k}(x_k) + \sum_{k=1}^K l_{e_k}(x_{k-1},x_k).
$$

\subsection{Shortest-walk problem}
The shortest $s\textsf{-}t$ walk in a GCS between points $\bar x_s$ and $\bar x_t$ is a walk of minimal cost.
It is the solution to the following optimization problem:
\begin{subequations}\label{e-swp-in-gcs}
\begin{align}
\inf_{K} \; \min_{w,\tau} \quad & l(w,\tau) &&\label{e-swp-objective} \\
\mathrm{s.t.} \quad & w\in\mathcal W^K_{s,t}, \quad \tau \in \mathcal T_w(\bar x_s, \bar x_t).
\end{align}
\end{subequations}
Note that we have a two-level optimization: we seek the shortest $K$-step walk $(w,\tau)$ at the inner level, and take the infimum over $K$ at the outer level.
We thus optimize over the discrete number of steps $K$, the walk $w$ through the graph, and the trajectory $\tau$ along this walk.

\begin{figure}[t!]
    \centering
    \begin{subfigure}[t]{\columnwidth}
        \includegraphics[width=0.9\columnwidth]{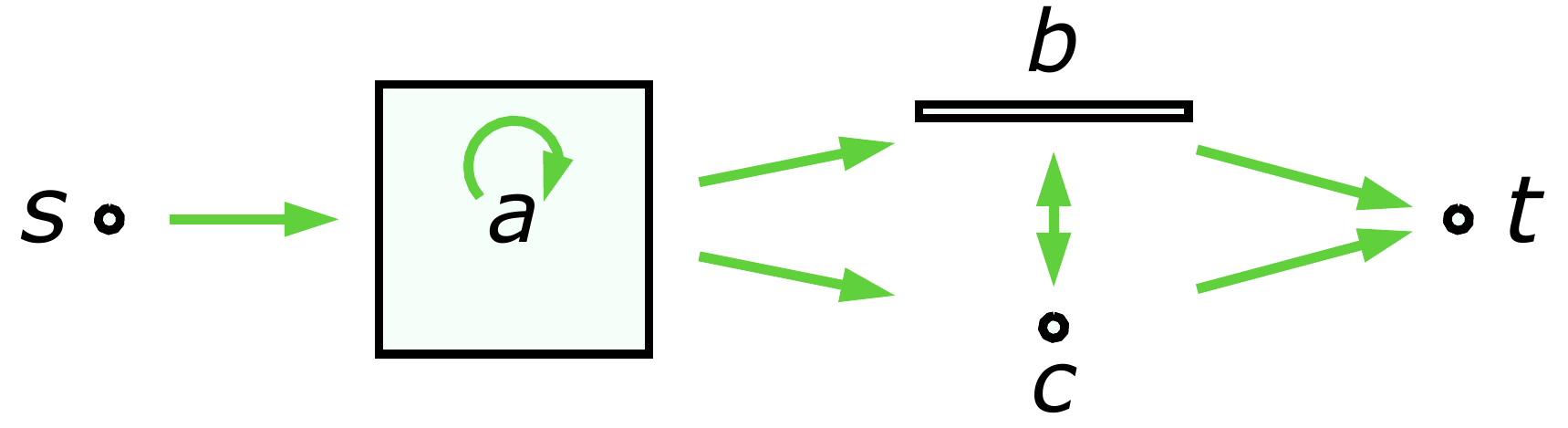}
        \caption{
        A GCS embedded in $\mathbb R^2$.
        }
        \label{sf-simple_graph}
    \end{subfigure}
    
    \begin{subfigure}[t]{\columnwidth}
        \includegraphics[width=0.9\columnwidth]{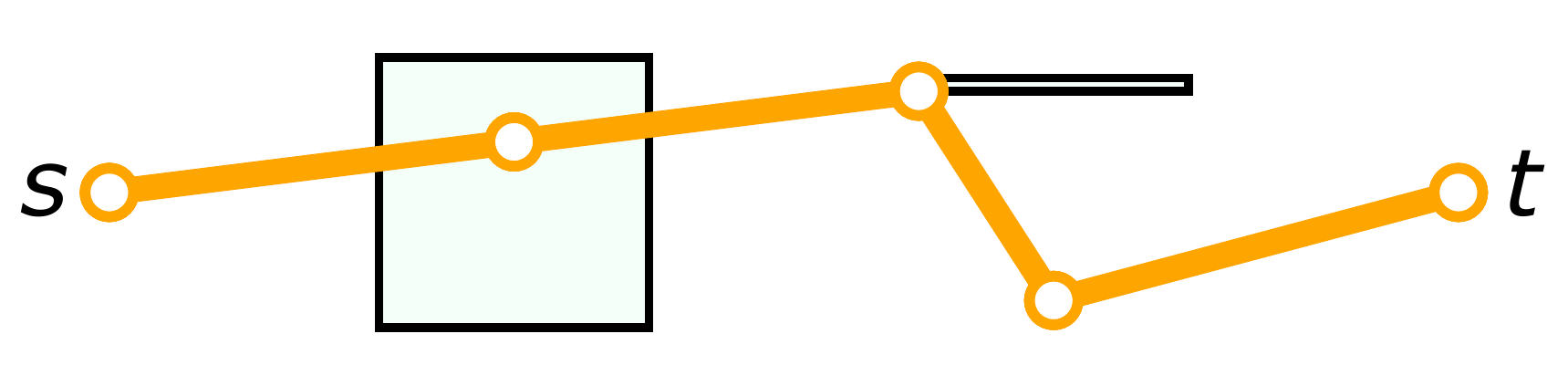}
        \caption{The shortest path between vertices $s$ and $t$ (orange) costs 35.}
        \label{sf-simple_path}
    \end{subfigure}
    
    \begin{subfigure}[t]{\columnwidth}
        \includegraphics[width=0.9\columnwidth]{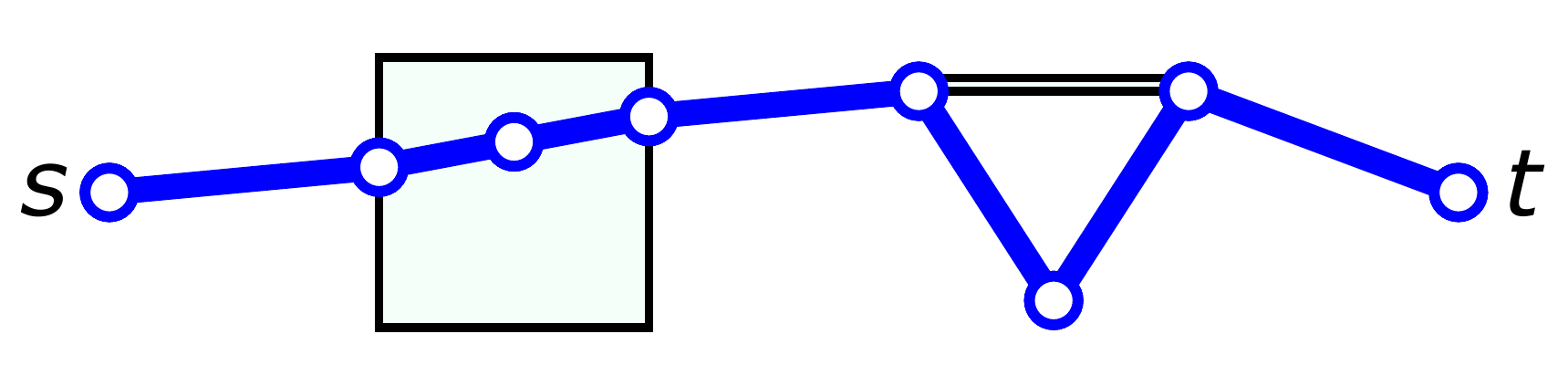}
        \caption{The shortest walk between vertices $s$ and $t$ (blue) costs 28. 
        Vertices $a$ and $b$ are both revisited along the walk: $a$ is visited three times consecutively, and $b$ is visited twice non-consecutively (note that unlike vertex $a$, there is no edge from vertex $b$ to itself).
        }
        \label{sf-simple_walk}
    \end{subfigure}

    \caption{
    A GCS where a shortest walk is not a path.
    }
    \label{f-simple_example}
\end{figure}

\subsection{Shortest walks need not be paths}
For an ordinary graph with non-negative edge costs, the shortest-walk problem always admits an optimal solution that is a path.
This is because revisiting a vertex creates a cycle of non-negative cost and makes no progress towards the target.
Therefore, this cycle can be removed without increasing the cost of the walk.
The same is not true for walks in a GCS.
When traversing a GCS, revisiting the same vertex can be advantageous, as demonstrated by the following example.

\begin{example}
Consider the 2D problem depicted in \Cref{sf-simple_graph}.
This GCS has 5 vertices $\mathcal V = \{s,a,b,c,t\}$ and 8 edges, drawn in green. 
The convex set $\mathcal X_a$ is a square, $\mathcal X_b$ is a segment, and $\mathcal X_s,\mathcal X_c,\mathcal X_t$ are points.
For every edge $e=(u,v)$, the edge cost is defined as $l_e(x_u,x_v) = 1 + ||x_u-x_v||_2^2$: the first term penalizes the number of steps taken, while the squared displacement term penalizes the size of each step.
There are no vertex costs, nor are there any additional edge constraints.
The solutions to the SPP and the SWP in this GCS are shown in \Cref{sf-simple_path,sf-simple_walk}.
To avoid revisiting vertices, the shortest path (orange) must take larger steps, incurring large penalties.
By taking smaller steps and revisiting vertices, the shortest walk (blue) achieves a lower cost.
Thus, even though cycles in the GCS still have a non-negative cost, they may help us make progress towards the target.
\end{example}

\subsection{Sufficient condition for finiteness of shortest walks}
\label{ss-sufficient-condition-for-finite-walk}
We note that the shortest walk in a GCS need not be of finite length $K$. 
Consider again the GCS in \Cref{sf-simple_graph}.
If the edge cost was $l_e(x_u,x_v) = ||x_u-x_v||_2^2$, then the optimal walk would involve taking infinitely many small steps through the vertex $a$.
As a result, no finite walk is optimal, though the solution does exist in the limit.
This issue is the reason for the infimum over the number of steps $K$ in program~\eqref{e-swp-in-gcs}.

The following is a simple sufficient condition for the optimal walk, if one exists, to be finite:
$$
\min \{l_e(x_u,x_v) \mid (x_u,x_v)\in \mathcal X_e\} >0,
$$
for every edge $e=(u,v) \in \mathcal E$.
This condition ensures that the cost of every step in the walk is bounded below by some positive value.
Thus an infinite walk must incur infinite cost, and cannot be optimal.
For practical purposes, this condition can be easily satisfied by adding a small $\epsilon>0$ to every edge cost $l_e$.
In what follows, we assume that this condition holds.

\section{Applications in robotics and control}
\label{s-many-problems-cast}

Depending on the application, searching for either paths or walks in a GCS can be a natural modeling choice. 
Paths are well-suited for problems where repeating behaviors is unnecessary or undesired, such as those involving unique actions, one-time traversals, or constrained resources. 
However, when such repetitions are necessary, as is often the case in robotics, walks provide a more natural and more general framework.
We now highlight several practical problems in robotics and optimal control where the shortest-walk formulation is particularly well-suited. 
We revisit these problems in \Cref{s-results} to provide experimental demonstrations.

\subsection{Collision-free motion planning with acceleration limits}
\label{ss-collision-free-planning-theory}

Motion planning around obstacles is a key challenge in robotics.
Sampling-based planners~\cite{kavraki1996probabilistic,lavalle1998rapidly,elbanhawi2014sampling} are a popular solution due to their simplicity; they offer probabilistic guarantees of feasibility and even optimality~\cite{karaman2011sampling,janson2015fast} but typically struggle with enforcing continuous dynamical constraints~\cite{webb2013kinodynamic,goretkin2013optimal,wu2020r3t}.
Optimization-based methods~\cite{betts1998survey,augugliaro2012generation,schulman2014motion,majumdar2017funnel,zhang2020optimization} effectively handle these constraints by framing them as part of a non-convex program, but often fail to find a feasible solution in complex environments due to their reliance on local solvers.
Planners based on mixed-integer optimization~\cite{schouwenaars2001mixed,richards2002aircraft,mellinger2012mixed,deits2015efficient} combine discrete and continuous search, using trajectory optimization for dynamics and branch-and-bound for global optimality.
The SPP in GCS is one such method~\cite{marcucci2023motion}, notable for fast solve times achieved through an effective convex formulation.
However, some desirable costs and constraints in this transcription remain non-convex.

\begin{figure}[t!]
    \centering
    \begin{subfigure}[t]{0.48\columnwidth}
        \centering
        \includegraphics[width=0.95\columnwidth]{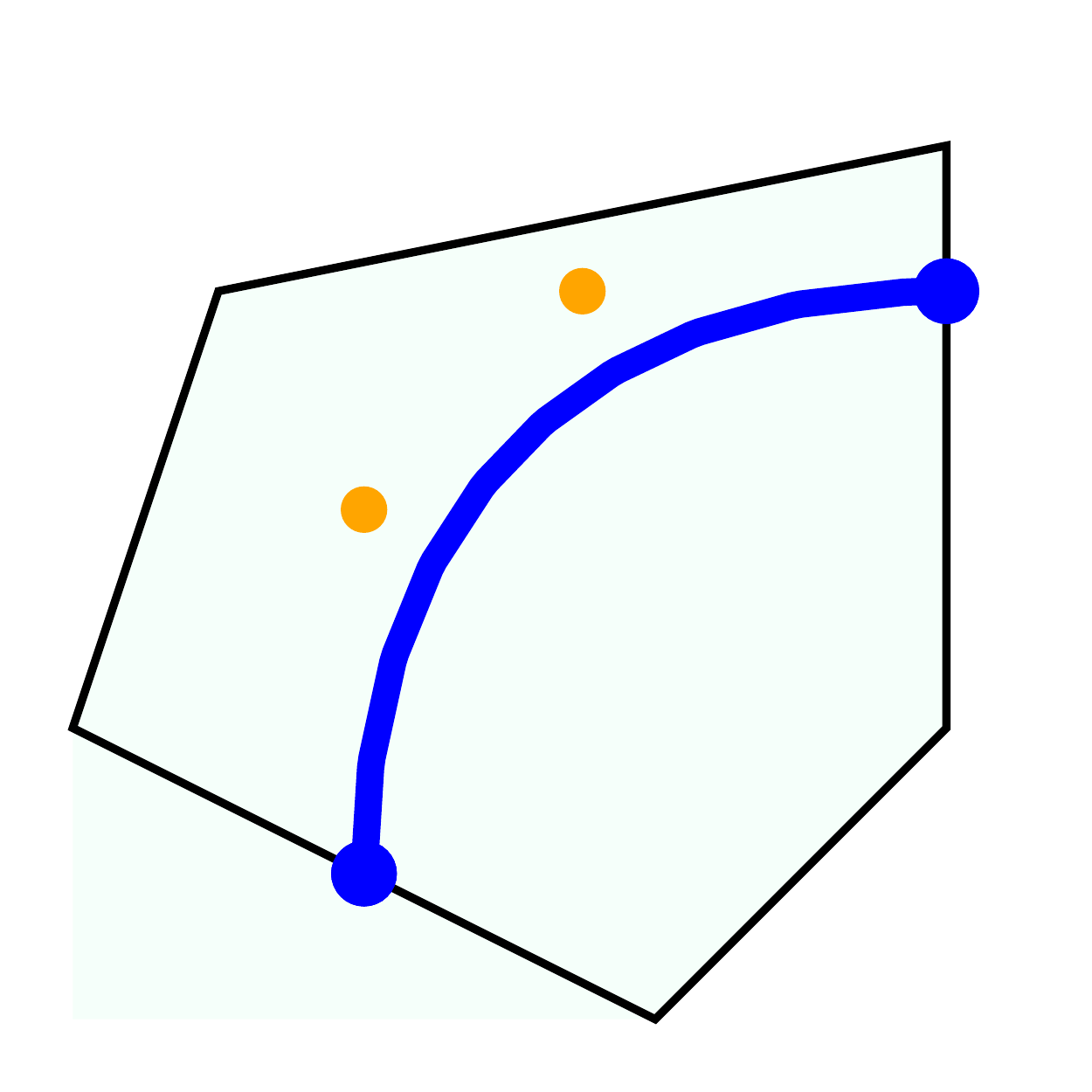}
        \caption{
        The shortest-path formulation in~\cite{marcucci2023motion} jointly searches for a Bézier curve and its duration per vertex, resulting in non-convex acceleration constraints.
        }
        \label{sf-bezier_spp_curves}
    \end{subfigure}\hfill
    \begin{subfigure}[t]{0.48\columnwidth}
        \centering
        \includegraphics[width=0.95\columnwidth]{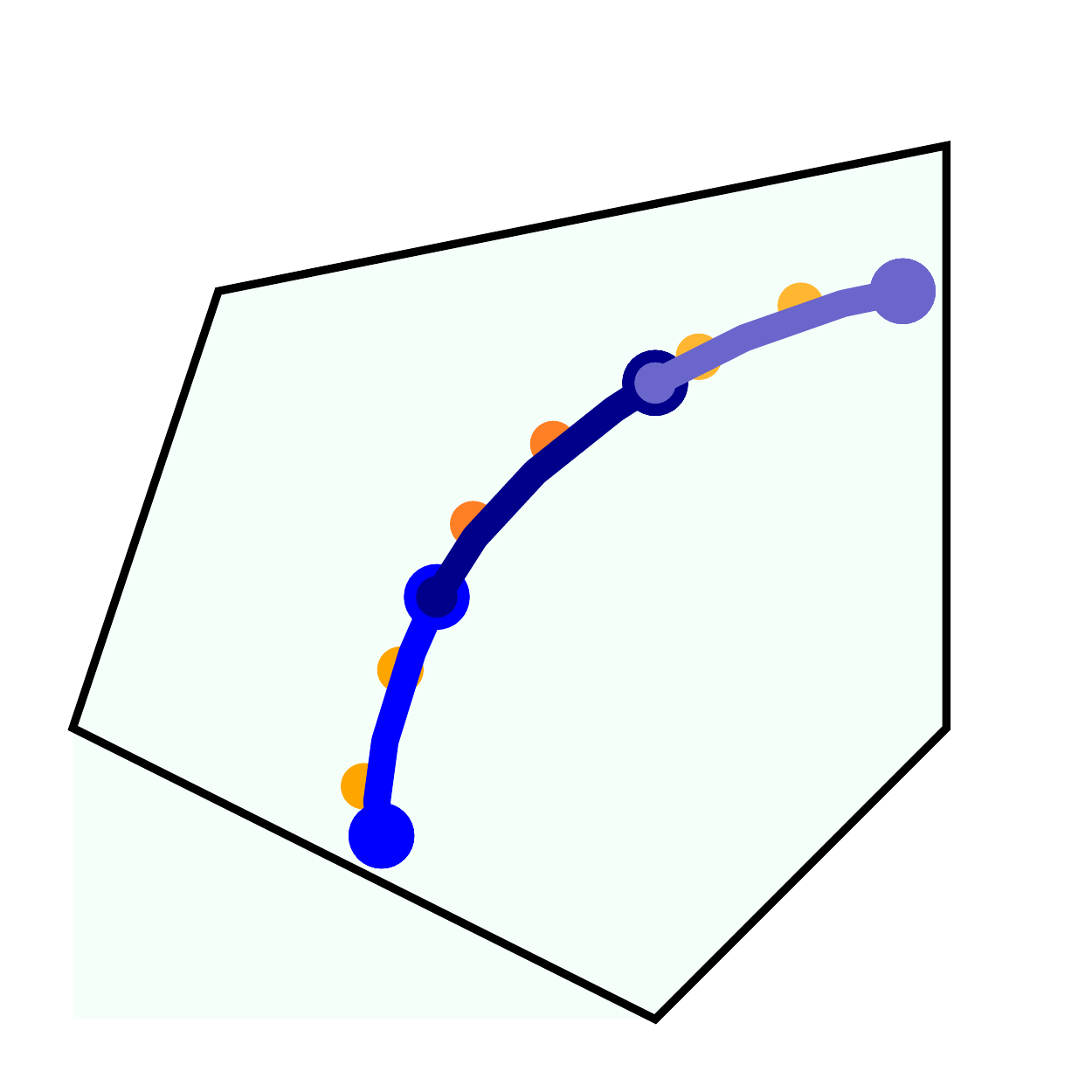}
        \caption{
        Our shortest-walk formulation searches for a fixed-duration Bézier curve per vertex visit, resulting in convex acceleration constraints.
        Above, the same vertex is revisited 3 times.
        }
        \label{sf-bezier_swp_curves}
    \end{subfigure}
    \caption{Contrasting the SPP and SWP formulations for the trajectory planning problem from \Cref{ss-collision-free-planning-theory}.
    Shown in blue are cubic Bézier curves, with their control points in orange.
    }
    \label{f-simple_bezier_example}
\end{figure}

To apply GCS to this problem, the authors in~\cite{marcucci2023motion} first decompose the collision-free space into polyhedral regions.
A GCS vertex is added for every collision-free region, and two vertices are connected by an edge if the corresponding regions overlap.
The convex set for each vertex is the set of Bézier curves contained within the corresponding region, together with the duration of that Bézier curve.
Visiting a vertex thus corresponds to selecting a trajectory through a collision-free region, pictured in \Cref{sf-bezier_spp_curves}; a vertex cost on the duration (and possibly other penalties) is also added.
Traversing an edge imposes additional constraints to guarantee a continuous and differentiable trajectory. 
A path in this GCS corresponds to a smooth, collision-free trajectory from start to target.

A key practical limitation of this formulation 
is that simultaneously optimizing the shape of the curve and its duration results in non-convex constraints on acceleration (and higher order derivatives).
To handle these higher-order derivative constraints, the trajectories must be post-processed using Time-Optimal Path Parameterization (TOPP)~\cite{verscheure2009time} or by solving a non-convex program on the vertex sequence~\cite{marcucci2024fast,von2024using}. 
However, this decouples the problem into separate stages, which can lead to suboptimal or infeasible solutions.

Alternatively, we can make the relevant constraints convex by fixing the duration of each curve.
To allow spending variable amount of time in each region, the shortest-path formulation requires vertex duplication; the time spent in each region becomes the product of the number of visited vertex duplicates and the duration of each curve.
This duration acts as a user-defined hyper-parameter, adjustable based on the application. 
However, this introduces a trade-off: longer durations result in large discretization errors, while smaller durations necessitate more vertex duplicates, increasing the size of the GCS instance and its computational complexity.

In contrast, the shortest-walk formulation naturally resolves this trade-off by allowing vertex revisits. 
Vertex duplication thus becomes unnecessary, and the problem’s complexity is not artificially inflated.
This effectively addresses the discretization trade-off, though arbitrarily small durations remain undesirable as they may result in excessively long optimal walks.
Shortest walks in GCS offer an alternative convex representation of the problem, yielding a more natural and compact formulation without artificially expanding the problem's description.

\subsection{Skill chaining}
\label{ss-behavior-composition-theory}

Consider a robotic system controlled by a discrete set of continuously parameterized skills (also referred to as motion primitives, actions, behaviors) that use low-level control policies to transition between configurations. 
Abstracting away the low-level dynamics of these policies, the goal of skill chaining is to select a sequence of skills and the corresponding control parameters that achieve the target state~\cite{sutton1999between,konidaris2009skill}. 
Related families of problems include sequential composition~\cite{burridge1999sequential,tedrake2009lqr} and Task and Motion Planning (see~\cite{garrett2021integrated} for a comprehensive review).
A common solution strategy alternates between sampling discrete skills and continuous transitions (control parameters), guided by strong heuristics~\cite{cambon2009hybrid,kaelbling2011hierarchical,srivastava2014combined,krontiris2016efficiently,garrett2020pddlstream}. 
However, to be effective in complex environments, these methods often rely on costly, hand-crafted samplers and may stall without them. 
To more effectively explore the space of continuous transitions and better inform discrete search, other approaches use optimization-based subroutines~\cite{toussaint2015logic,hadfield2016sequential,shoukry2018smc,fernandez2018scottyactivity}.
The GCS-based formulation presented below is in this vein.

Given an $n$-dimensional configuration space, we define each skill $\pi$ via a set ${\mathcal Q_\pi\subset\mathbb R^{2n}}$ of feasible configuration transitions ${(q,q')\in \mathcal Q_{\pi}}$ that can be achieved by this skill.
Note that alternative definitions in the literature describe skills through preconditions (pre-image, domain, initiation set) and effects (reachable or goal sets, termination condition), but all are generally interchangeable.
Each skill also has an associated cost function $c_{\pi}: \mathcal{Q}_{\pi} \rightarrow \mathbb{R}_+$, where ${c_{\pi}(q, q')}$ is the cost of the transition from configuration $q$ to $q'$ under this skill.
Given a pair of start and target configurations $\bar q_s, \bar q_t$, the goal is to find a sequence of skills $(\pi_1,\ldots, \pi_K)$ and the sequence of transitions $\big((q_0,q_1),\ldots, (q_{K-1}, q_K)\big)$, such that $q_0=\bar q_s$, $q_K=\bar q_t$, and each transition $(q_{k-1},q_k)$ is achieved via $\pi_k$.

To use GCS, we require the sets $\mathcal Q_{\pi}$ and the cost functions $c_\pi$ to be convex.
If they are not, we assume that convex approximations or decompositions are available (though these may be difficult to obtain, existing tools in the GCS ecosystem can help~\cite{deits2015computing,petersen2023growing,werner2024faster}).
In our GCS formulation, each skill $\pi$ corresponds to a vertex with a convex set $\mathcal{X}_{\pi} = \mathcal Q_\pi$ and a vertex cost $l_\pi = c_\pi$.
Visiting a vertex that correspond to $\pi$ is thus equivalent to executing a transition $(q,q')$ and incurs a vertex cost $c_\pi(q,q')$.
An edge connects two vertices if their skills can be chained: that is, if there exist configurations $q_0,q_1,q_2$ such that $(q_0, q_1) \in \mathcal{Q}_{\pi_1}$ and $(q_1, q_2) \in \mathcal{Q}_{\pi_2}$. 
Ensuring that the end point $q_1$ of the first skill is also the start point of the second skill requires adding an appropriate edge constraint.
We then add a start vertex for the start configuration $\bar{q}_s$ and connect it to vertices that represent skills executable from $\bar{q}_s$.
We add a target vertex in a similar fashion.
The shortest walk in this GCS is exactly the solution to the skill chaining problem: it is a sequence of skills $(\pi_1,\ldots, \pi_K)$ together with the corresponding sequence of transitions $\big((q_0,q_1),\ldots, (q_{K-1}, q_K)\big)$.

\subsection{Optimal control for hybrid systems}
\label{ss-pwa-hybrid-dynamics-theory}

Many challenging problems in robotics, such as footstep planning, planning through contact, and dexterous manipulation, involve systems with hybrid dynamics.
It is well known that such systems can be approximated arbitrarily-well with a Piecewise Affine (PWA) dynamical model~\cite{sontag1981nonlinear,sontag1995interconnected,bemporad1999control}.
Motivated by this, we consider the problem of optimal control for discrete-time PWA dynamical systems.
We refer the reader to~\cite{lee2024strong} for a recent review of hybrid system control approaches, which highlights the SPP in GCS as an effective and competitive strategy.
Below we show that the shortest-walk formulation may offer a more effective strategy.

\newcommand{\myconst}{3cm}
\newcommand{\pendulumStatespaceBase}[1][black]{
    \pgfmathsetmacro{\axislength}{4pt} 
    \pgfmathsetmacro{\axislengthhalf}{0.5*\axislength} 

    \coordinate (origin) at (0,0);
    \coordinate (x_end) at (\axislength, 0);
    \coordinate (y_pos) at (0, \axislengthhalf);
    \coordinate (y_neg) at (0, -\axislengthhalf);

    \fill[blue!20, draw = black!50] (y_neg) -- (y_pos) -- ($0.5*(x_end)+(y_pos)$) -- ($0.5*(x_end)+(y_neg)$) -- cycle;
    \fill[green!20, draw = black!50] ($0.5*(x_end)+(y_pos)$) -- ($0.5*(x_end)+(y_neg)$) -- ($(x_end) + (y_neg)$) -- ($(x_end) + (y_pos)$) -- cycle;

    \draw[very thick, ->, #1] (origin) -- ($(x_end)+(0.2,0)$) node[right] {$\theta$}; 
    \draw[very thick, ->, #1] (origin) -- ($(y_pos)+(0,0.2)$) node[above] {$\dot{\theta}$};
    \draw[very thick, -, #1] (origin) -- ($(y_neg)-(0,0.2)$) {};
}
\newsavebox{\pendulumGraphPartition}
\sbox{\pendulumGraphPartition}{
\begin{tikzpicture}
    \pendulumStatespaceBase[opacity=0]
    \pgfmathsetmacro \nx{1/4*\axislength};
    \pgfmathsetmacro \ny{0};
    \pgfmathsetmacro \cx{3/4*\axislength};
    \pgfmathsetmacro \cy{0};

    \tikzset{every path/.style={line width=0.5mm}}
    \node (N) at (\nx, \ny) [circle, draw] {$N$};
    \node (C) at (\cx, \cy)[circle, draw] {$C$}; 

    \draw[->] (N) edge[loop left] (N);
    \draw[->] (C) edge[loop right] (C); 
    
    \draw[->] (N) to[bend left] (C);
    \draw[->] (C) to[bend left] (N);
\end{tikzpicture}
}

\newsavebox{\pendulumTrajectory}
\sbox{\pendulumTrajectory}{
\begin{tikzpicture}[font=\large] 
    \pendulumStatespaceBase

    \tikzset{trajnode/.style={draw=blue, fill=yellow!20, circle, thick, inner sep = 2pt}}

    \node[trajnode, label={left:$s_{0}$}] (xs) at ($(0.7,0) + 0.5*(y_pos)$) {};
    \node[trajnode, label=right:$s_{2}$] (x2) at ($(y_pos) + 0.5*(x_end) - (0, 0.3)$) {};
    \node[trajnode, label=below right:$s_{5}$] (x5) at ($0.5*(x_end)+ 0.6*(y_neg)$) {};
    \node[trajnode, label={[label distance=0.75mm]273:$s_{t}$}](xe) at (origin) {};

    \tikzset{edge/.style={draw=blue, line width=0.75mm}}
   
    \draw[edge] (xs) .. controls ($(xs)+(0.5, 0.5)$) .. (x2)
    node[midway, trajnode, label=left:$s_{1}$] {};

    \draw[edge] (x2) .. controls ($0.5*(x2)+0.5*(x5)+(0.9,0.5)$) and ($0.25*(x2)+0.75*(x5)+(1.12,0.5)$).. (x5)
    node[pos=0.3, trajnode, label=right:$s_{3}$] {}
    node[pos=0.75, trajnode, label=right:$s_{4}$] {}
    ;

    \draw[edge] (x5) .. controls ($(x5)-(1, -0.4)$) .. (xe)
    node[midway, trajnode, label=below:$s_{6}$] {};
    
\end{tikzpicture}
}

\begin{figure}[t!]
    \centering
    \begin{subfigure}[b]{0.27\columnwidth}
        \centering
        \resizebox{\linewidth}{!}{\begin{circuitikz}[scale=1]
    \coordinate (origin) at (0,0);
    \pgfmathsetmacro{\pendlength}{3.3pt} 
    \pgfmathsetmacro{\angle}{15} %
    \pgfmathsetmacro{\pendx}{\pendlength*sin(\angle)}
    \pgfmathsetmacro{\pendy}{\pendlength*cos(\angle)}

    \coordinate (pend_head) at (\pendx, \pendy);
    \draw[thick] (origin) -- (pend_head); 
    \fill[black] (pend_head) circle (6pt); 
    \fill[black] (origin) circle (1pt); 
    \draw[dashed] (origin) -- (0, \pendlength);  

    \pgfmathsetmacro{\arcradius}{\pendlength/2}
    \pgfmathsetmacro{\arcstartx}{\arcradius*sin(\angle)}
    \pgfmathsetmacro{\arcstarty}{\arcradius*cos(\angle)}
    \pgfmathsetmacro{\startangle}{90-\angle}; %
    \pgfmathsetmacro{\endangle}{90}; %
    \draw[thick, dashed] ([shift=(origin)] \startangle:\arcradius) arc[start angle=\startangle, end angle=\endangle, radius=\arcradius];
    \pgfmathsetmacro{\arcmiddlex}{1.2*\arcradius*sin(\angle/2)}
    \pgfmathsetmacro{\arcmiddley}{1.2*\arcradius*cos(\angle/2)}
    \node at (\arcmiddlex, \arcmiddley) {$\theta$};

    \pgfmathsetmacro{\softwalldist}{1.5pt}
    \coordinate (softwallbottom) at (\softwalldist, 0);
    \coordinate (softwalltop) at (\softwalldist, \pendlength);
    \draw[very thick] (softwallbottom) -- (softwalltop);

    \pgfmathsetmacro{\hardwallstartx}{\softwalldist+1}
    \pgfmathsetmacro{\hardwallendx}{\hardwallstartx+0.5}
    \coordinate (hardwallbottomleft) at (\hardwallstartx, 0);
    \coordinate (hardwallbottomright) at (\hardwallendx, 0);
    \coordinate (hardwalltopright) at (\hardwallendx, \pendlength);
    \coordinate (hardwalltopleft) at (\hardwallstartx, \pendlength);
    \draw[EDB] (hardwallbottomleft) -- (hardwallbottomright) -- (hardwalltopright) -- (hardwalltopleft);

    \coordinate (softwallmiddle) at (\softwalldist, \pendlength/2);
    \coordinate (hardwallmiddleleft) at (\hardwallstartx, \pendlength/2);
    \draw (softwallmiddle) to [spring] (hardwallmiddleleft); 

\end{circuitikz}}
        \caption{}
        \label{sf-pwa-1}
    \end{subfigure}\hfill
    \begin{subfigure}[b]{0.34\columnwidth}
        \centering
        \resizebox{\linewidth}{!}{\usebox{\pendulumGraphPartition}}
        \caption{}
        \label{sf-pwa-2}
    \end{subfigure}\hfill
    \begin{subfigure}[b]{0.34\columnwidth}
        \centering
        \resizebox{\linewidth}{!}{\usebox{\pendulumTrajectory}}
        \caption{}
        \label{sf-pwa-3}
    \end{subfigure}
    \caption{An actuated pendulum with a soft wall (a) can be approximated as a PWA system with two modes (no-contact and contact), which can be modeled as a GCS with two vertices and four edges (b). 
    An optimal state-space trajectory for regulating the pendulum to the equilibrium position $s_t=(0,0)$ can be computed by solving a SWP in GCS, shown in (c).
    }
    \label{f-pwa-example}
\end{figure}
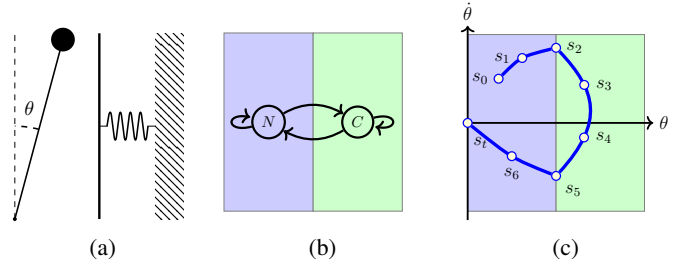

Let $\mathcal S$ and $\mathcal A$ be our system's state and control spaces, and let the state-space be partitioned into closed, polyhedral sets $\mathcal S = \cup_i \mathcal S_i$, commonly referred to as modes.
A PWA control system evolves evolves according to different affine dynamics depending on the mode that system is in.
That is, the system's dynamics at time-step $n$ are governed by:
\begin{align*}
    s_{n+1} = A_{i}s_{n} + B_{i}a_{n} + c_{i}, \text{ if } s_n \in \mathcal {S}_{i},\; a_n\in\mathcal A.
\end{align*}
Executing control input $a_n$ at state $s_n\in\mathcal S_i$ of mode $i$ incurs the mode-specific stage cost $l_i(s_n, a_n)$.
The PWA optimal control problem seeks a state, control, and mode trajectories between source and target states $\bar s_0$ and $\bar s_t$, satisfying the PWA dynamics and minimizing the total stage cost.

This problem can be naturally cast as a shortest walk in a GCS. 
We illustrate this in \Cref{f-pwa-example} for a pendulum with a soft wall, a canonical benchmark for control through contact that can be effectively approximated as a PWA system.
For every mode $i$, we define a GCS vertex $i$ with a convex set $\mathcal X_i = \mathcal S_i\times\mathcal A$.
Two vertices are connected with an edge if a feasible transition exists between some pair of states in the corresponding modes.
Affine dynamics are imposed as edge constraints, and the convex stage cost $l_i$ is added as a vertex cost.
For the pendulum, this results in a GCS with two vertices ($C$ for contact, $N$ for no-contact) and four edges, pictured in \Cref{sf-pwa-2}.
Note that the affine dynamics along the edges $(N,C)$ and $(C,N)$ are different.
The shortest walk in this GCS is a vertex sequence $w$, which corresponds to a PWA mode trajectory, and a sequence of points $\tau$, which corresponds to state and control trajectories. 
This is illustrated in \Cref{sf-pwa-3}.

\section{Solution method}
\label{s-solution-method}

Similar to the SPP in GCS, the SWP in GCS is also NP-hard.
The authors in~\cite[\S 9.2]{marcucci2024graphs} reduced the well-known NP-complete 3SAT problem~\cite{karp2010reducibility} to the shortest path in an acyclic GCS.
This reduction proves that the shortest walks are NP-hard as well. 
Indeed, since every walk is a path in an acyclic graph, the SPP and the SWP in an acyclic GCS are equivalent, and the NP-hardness of the SWP follows.
Therefore, we do not expect to find an efficient solution for each problem instance.

\begin{figure}[t!]
    \centering
    \begin{subfigure}[t]{0.28\columnwidth}
        \centering
        \includegraphics[width=0.95\columnwidth]{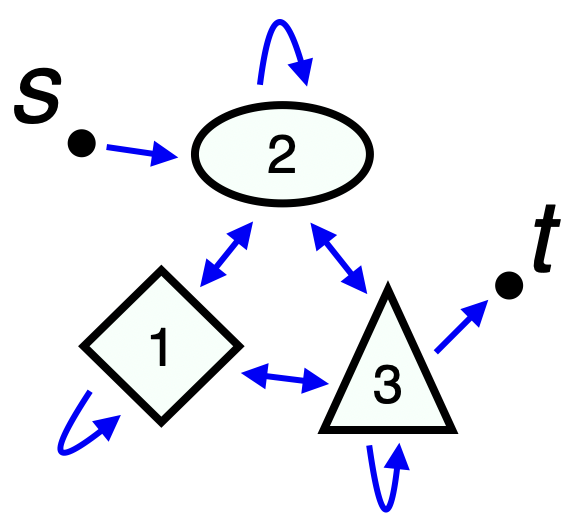}
        \caption{
        }
        \label{sf-walk-graph}
    \end{subfigure}\hfill
    \begin{subfigure}[t]{0.71\columnwidth}
        \centering
        \includegraphics[width=0.95\columnwidth]{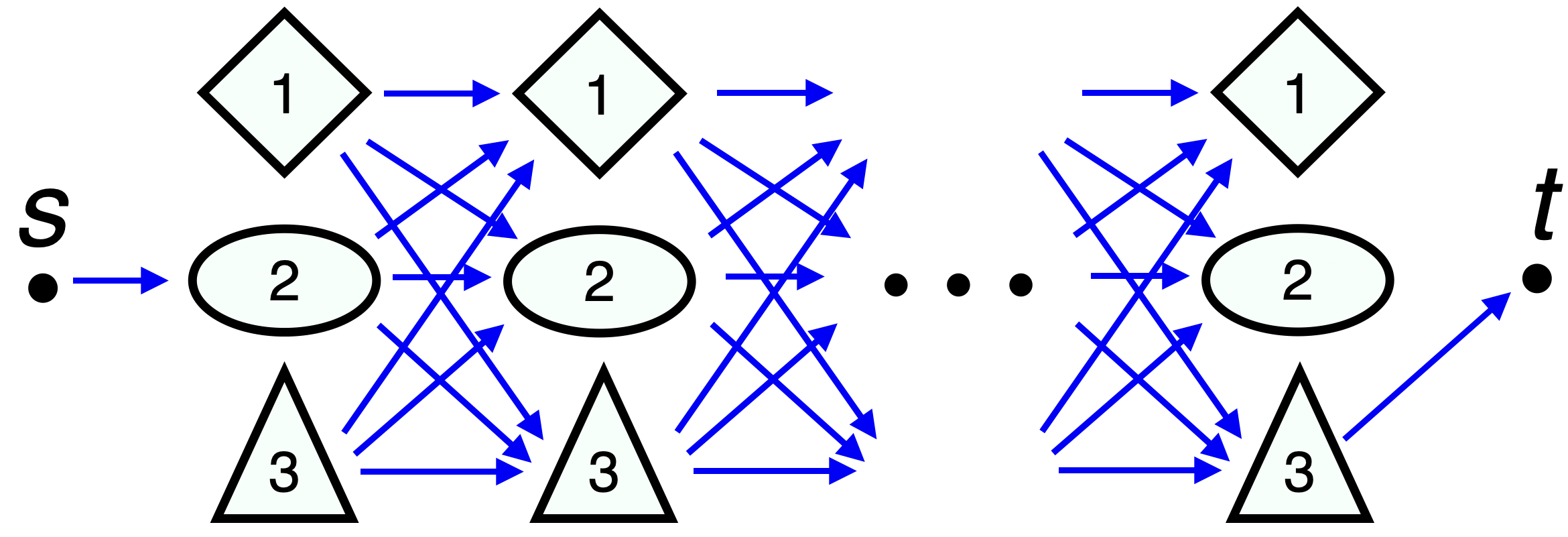}
        \caption{
        }
        \label{sf-path-graph}
    \end{subfigure}
    \caption{
    A naive approach to solving the SWP in GCS in (a) involves constructing a layered GCS in (b) by duplicating vertices across $K-1$ layers, for each $K \to \infty$.
    In contrast, our approach operates directly on the original, smaller GCS in (a), and avoids artificially increasing the size of the problem. 
    }
    \label{f-path-and-walk-graph-comparison}
\end{figure}

A naive way to find the shortest walk in a GCS is to compute $K$-step optimal walks for progressively larger $K$ and maintain the best solution.
As $K\to\infty$, this converges to the infimum in \eqref{e-swp-in-gcs}.
A $K$-step optimal walk can be obtained by solving an SPP in a different GCS, where we duplicate vertices as a way to allow revisits, as shown in \Cref{f-path-and-walk-graph-comparison}.
Given the original GCS in \Cref{sf-walk-graph}, we construct a new layered GCS in \Cref{sf-path-graph}, with each layer containing duplicates of the original vertices, and consecutive layers connected based on the original edges.
Solving the SPP in this layered GCS yields a $K$-step optimal path, equivalent to a $K$-step optimal walk in the original GCS.
This layered construction was considered~\cite[\S 10.2.3]{marcucci2024graphs}, where it was shown to be computationally expensive. 
Moreover, for the shortest walks, this approach is intractable as it requires solving SPP queries over increasingly larger GCS instances.

Incremental search offers an effective approach to solving the problem. 
Recent works proposed incremental methods for solving the SPP in GCS~\cite{chia2024gcs,sundar2024graphs,natarajan2024implicit,morozov2024multi}.
These techniques naturally extend to produce walks by explicitly allowing vertex revisits.
The performance of incremental search depends significantly on the quality of the guiding heuristic.
Leveraging the methodology from~\cite{morozov2024multi}, we develop effective heuristics for the shortest walks in GCS.

First, we derive the Bellman equation for this problem, characterizing the shortest-walk cost-to-go function at each GCS vertex.
Next, we solve this Bellman equation, obtaining the exact cost-to-go function as the solution to an infinite-dimensional optimization problem. 
To get a finite-dimensional numerical approximation, we use semidefinite programming, searching instead for quadratic lower bounds on the cost-to-go at each vertex.
We emphasize that this program is finite and scales with the size of the GCS --- not the length of the shortest walk, which can be arbitrarily long.
Using these lower bounds, we apply incremental search to extract a walk.

\subsection{Bellman equation}
\label{ss-bellman-equation}
We observe that the \textit{principle of optimality} holds for the SWP in GCS, stating that every subwalk of a shortest walk $(w,\tau)$ is also a shortest walk.
Indeed, if a subwalk was not itself optimal, then it could be replaced with the actual optimal subwalk, resulting in a walk of lower cost than the original shortest walk: a contradiction.
Leveraging this property, we derive the Bellman equation for the SWP in GCS, which characterizes the cost-to-go function for this problem.

For every vertex $v\in\mathcal V$ and point $x_v\in\mathcal X_v$, let $J_v^*(x_v)$ denote the cost of a shortest walk from $x_v$ to the target point $\bar x_t$, also referred to as the \textit{cost-to-go} function.
Consider a point $x_u$ of vertex $u$, and let the point $x_v$ of vertex $v$ be next along a shortest walk from $x_u$ (where vertices $u,v$ need not be distinct, as this is a walk).
Then by the principle of optimality, the cost-to-go $J_u^*(x_u)$ must be the sum of the incurred costs $l_u(x_u) + l_{(u,v)}(x_u,x_v)$ and the subsequent cost-to-go $J_{v}^*(x_v)$.
Furthermore, since the transition to $x_v$ of vertex $v$ is optimal, it must minimize this sum among all other feasible transitions.
This is summarized in the Bellman equation below:
\begin{equation}\label{e-bellman-equation}
\begin{aligned}
J_{u}^*(x_u) \;\;=\;\; \min_{x_v,\, v} & \quad l_u(x_u) + l_e(x_u, x_v) + J_{v}^*(x_v) \\
\text{s.t.} &\quad  e=(u,v)\in\mathcal E, \\
& \quad (x_u,x_v)\in\mathcal X_e.
\end{aligned}
\end{equation}

\subsection{Synthesis of cost-to-go lower bounds}
\label{ss-synthesis-lower-bounds}

We now formulate an optimization problem that searches for the cost-to-go function that solves the Bellman equation~\eqref{e-bellman-equation}.
It is an infinite-dimensional Linear Program (LP); we discuss a tractable numerical approximation in \Cref{ss-sdp}.

We relax the Bellman equation~\eqref{e-bellman-equation} and search over the functions $J_v$ that satisfy the following inequality:
\begin{align}
&J_u(x_u) \leq l_u(x_u) + l_e(x_u, x_v)  + J_v(x_v), 
\nonumber
\end{align}
for every edge $e=(u,v)\in\mathcal E$ and for every feasible pair of points $(x_u,x_v)\in \mathcal X_e$.
This inequality states that $J_u(x_u)$ is no higher than the incurred vertex and edge costs $l_u(x_u) + l_e(x_u,x_v)$ plus the subsequent value $J_v(x_v)$.
Constraining $J_t(\bar x_t) =  l_t(\bar x_t)$ at the target, the resulting functions $J_u$ must be lower bounds on the cost-to-go $J_u'$, that is: $J_u(x_u)\leq J_u'(x_u)$ for all points $x_u\in\mathcal X_u$ and vertices $u\in\mathcal V$.
To make the function $J_s$ a tight lower bound on the cost-to-go $J_s^*$ at the source point $\bar x_s$, we maximize the value $J_s(\bar x_s)$. 
We obtain the following program: \begin{subequations}
\label{e-walk-cost-to-go-synthesis}
\begin{align}
\max  \quad & J_s(\bar x_s) \label{e-w-a} \\
\text{s.t.} \quad & J_v: \mathcal X_v\rightarrow\mathbb R_+,  \hspace{3.05cm} \forall v\in \mathcal V,\label{e-w-b}\\
&  J_u(x_u) \leq  l_u(x_u) + l_e(x_u, x_v) + J_v(x_v), \label{e-w-c}\\
& \hspace{5.3cm} \forall e=(u,v)\in\mathcal E,\nonumber\\
& \hspace{5.3cm}  \forall (x_u,x_v)\in\mathcal X_e, \nonumber\\
& J_t(\bar x_t) = l_t(\bar x_t). \label{e-w-d}
\end{align}
\end{subequations}
Constraints \eqref{e-w-c} and \eqref{e-w-d} enforce that $J_v$ is a lower bound on $J_v^*$ for every vertex $v \in \mathcal{V}$.
The objective function \eqref{e-w-a} maximizes the value of the lower bound $J_s(\bar x_s)$, which maximized when the lower bound is tight: $J_s(\bar x_s) = J_s^*(\bar x_s)$.
Thus the optimal solution to program \eqref{e-walk-cost-to-go-synthesis} yields an exact solution to the Bellman equation~\eqref{e-bellman-equation} at the source point $\bar x_s$.

Simultaneously maximizing $J_v(x_v)$ across all ${x_v\in\mathcal X_v}$ and vertices $v\in\mathcal V$ yields tight lower bounds on the cost-to-go over the entire GCS, akin to the classical many-to-one SPP.
Furthermore, similar to the classical many-to-many SPP, program \eqref{e-walk-cost-to-go-synthesis} can be generalized to produce lower-bounds on a cost-to-go $J^*_{v,t}(x_v,x_t)$ that is also a function of the target state $x_t$.
For further details, see~\cite[App. A]{morozov2024multi}, which explores the many-to-many generalization of the SPP in GCS.
This cost-to-go function is extremely useful, as it can be reused for multiple shortest-walk queries over the same GCS.

Program \eqref{e-walk-cost-to-go-synthesis} is an infinite-dimensional LP, as both the constraints and the objective are linear.
To obtain an approximate numerical solution, we use a finite-dimensional formulation, naturally sacrificing the tightness of the lower bounds.

\subsection{Tractable finite-dimensional cost-to-go lower bounds}
\label{ss-sdp}
We employ Semidefinite Programming (SDP) to produce an approximate numerical solution to \eqref{e-walk-cost-to-go-synthesis}.
We briefly outline the approach and direct the reader to~\cite[\S 3.2]{morozov2024multi} for further details.

The key challenge in program \eqref{e-walk-cost-to-go-synthesis} lies in enforcing the non-negativity constraints \eqref{e-w-b}, \eqref{e-w-c}: it is generally hard to numerically impose the constraint that a function is non-negative over a continuous set of points.
However, if we restrict the function to be polynomial and restrict the set to be basic semi-algebraic, then this constraint can be enforced in a convex manner via a Linear Matrix Inequality~\cite[\S 3.2.4]{blekherman2012semidefinite}.
We restrict the lower bounds $J_v$ per vertex $v$ to be convex quadratic, thus searching over the coefficients of the quadratic polynomials $J_v$.
Additionally, we restrict convex sets $\mathcal X_v$ to be intersections of polyhedra and ellipsoids, and restrict the cost functions $l_v, l_e$ to be convex quadratics (using quadratic approximations for non-quadratic $l_v, l_e$).
These restrictions allow us to cast program~\eqref{e-walk-cost-to-go-synthesis} as a tractable SDP, producing quadratic lower bounds on the cost-to-go function at each vertex.
While arbitrary-degree polynomial lower bounds can also be produced via the 
Sums-of-Squares hierarchy~\cite{parrilo2000structured,parrilo2003semidefinite,lasserre2001global}, using quadratic polynomials strikes a balance between expressive power and computational complexity.

This cost-to-go synthesis step is the most computationally intensive part of our solution; still, it is surprisingly tractable in practice, taking seconds to a minute even for complex systems and environments.
Additionally, this computational effort can be justified in a multi-query setting, where the computation only needs to be performed once and can then be reused for all subsequent shortest-walk queries over the same GCS.
This is particularly advantageous in multi-query robotic applications where a robot must repeatedly solve similar planning tasks within a fixed environment.

\subsection{Extracting walks from the approximate cost-to-go}
\label{ss-incremental-search}

We extract an approximate solution to the SWP in GCS using incremental search, similar to~\cite{morozov2024multi,chia2024gcs}, using cost-to-go lower bounds as a guiding heuristic.
Specifically, we use multi-step lookahead greedy search, as it quickly produces effective heuristic solutions.
We briefly describe the strategy below.

The solution is constructed incrementally, one vertex at the time.
At iteration $k$ of the search, let $(v_k,x_k)$ be the current vertex and vertex point (initialized with $(s, \bar x_s)$), and let $n$ be the lookahead horizon.
We consider all candidate $n$-step decision sequences $(w,\tau)$ that originate at $(v_k,x_k)$ and, among them, greedily select the one that minimizes the $n$-step lookahead cost-to-go:
\begin{subequations}
\label{e-policy-spp-gcs}
\begin{align}
\underset{ (w,\,\tau)}{\min} & \quad  l(w,\tau) + J_{v_{k+n}}(x_{k+n}) - l_{v_{k+n}}(x_{k+n}) \label{e-policy-objective} \\
\text{s.t.} & \quad w \in\mathcal W^n_{v_k}, \label{e-policy-searching-for-path} \\
& \quad \tau \in\mathcal T_w(x_k, x_{k+n}). \label{e-policy-searching-for-trajectory}
\end{align}
\end{subequations}
In~\eqref{e-policy-searching-for-path}, we consider all $n$-step walks $w$ that originate at vertex $v_k$, described by the set $\mathcal{W}^n_{v_k}$.
We define this set to also include walks that terminate at the target early: walks that start at $v_k$, end at $t$, and have fewer than $n$ steps.
In~\eqref{e-policy-searching-for-trajectory}, for each candidate walk $w$, we consider a trajectory ${\tau}$ along this walk that starts at the fixed point $x_k$ and ends at a variable point $x_{k+n}$.
If the candidate walk $w$ ends early on the target~$t$, we modify the constraint~\eqref{e-policy-searching-for-trajectory} to be $\tau \in\mathcal T_w(x_k, \bar x_t)$, so that the trajectory $\tau$ ends on the target point.
Together, $(w,\tau)$ is a candidate $n$-step decision sequence in the GCS.
The objective~\eqref{e-policy-objective} minimizes the $n$-step lookahead cost-to-go: the cost of the $n$-step lookahead sequence $l(w,\tau)$, plus the remaining cost-to-go lower bound $J_{v_{k+n}}(x_{k+n})$, minus the last vertex cost $l_{v_{k+n}}(x_{k+n})$ to avoid double-counting.

\begin{figure*}[t!]
    \centering
    \begin{tabular}{ccc}
        \begin{subfigure}[t]{0.31\textwidth}
            \centering 
            \includegraphics[width=\textwidth]{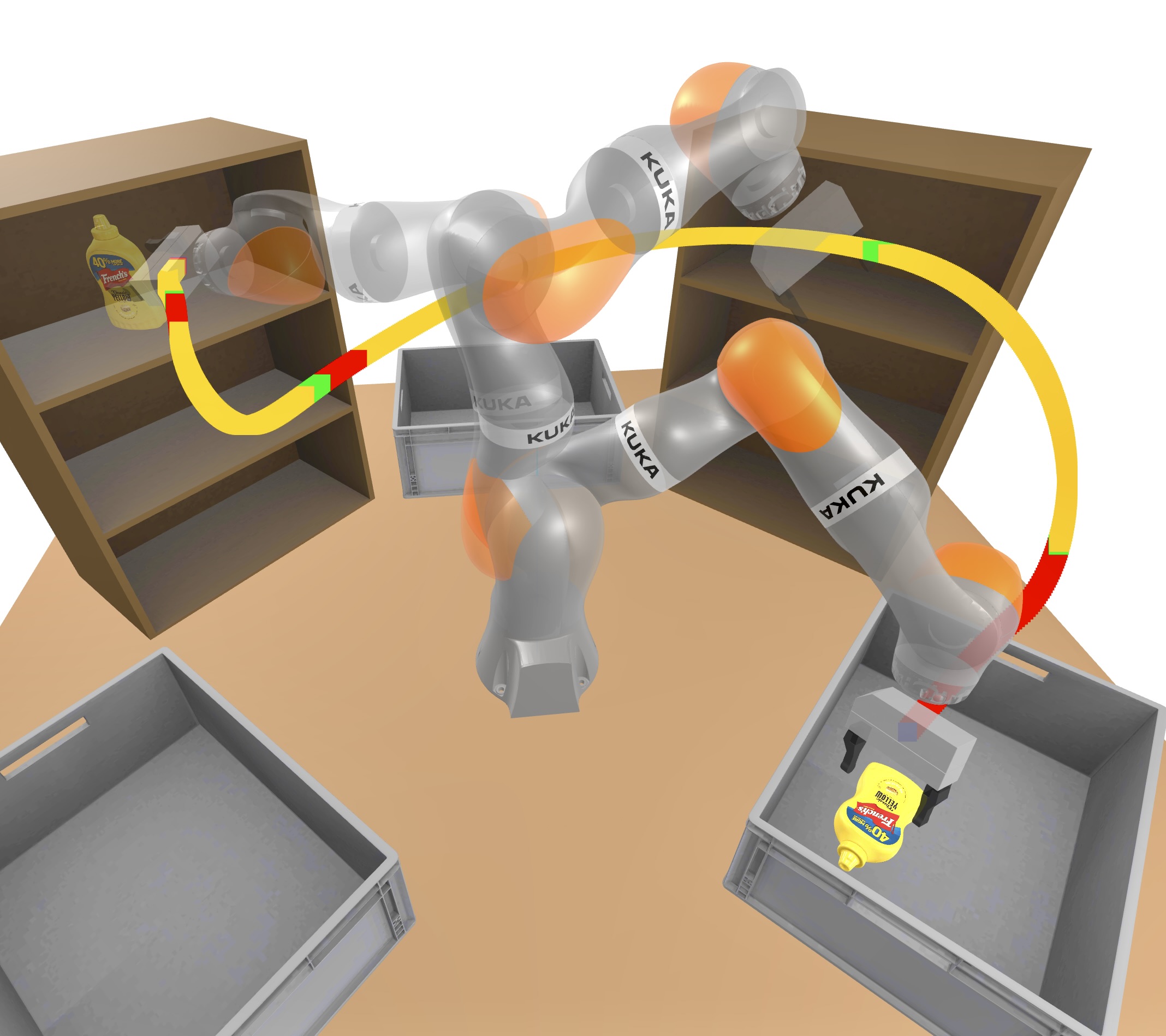}
            \caption{SPP in GCS
            generates a velocity-limited trajectory of duration 2.21s, which violates acceleration limits (red).}
            \label{sf-arm-spp}
        \end{subfigure} &
        \begin{subfigure}[t]{0.31\textwidth}
            \centering 
            \includegraphics[width=\textwidth]{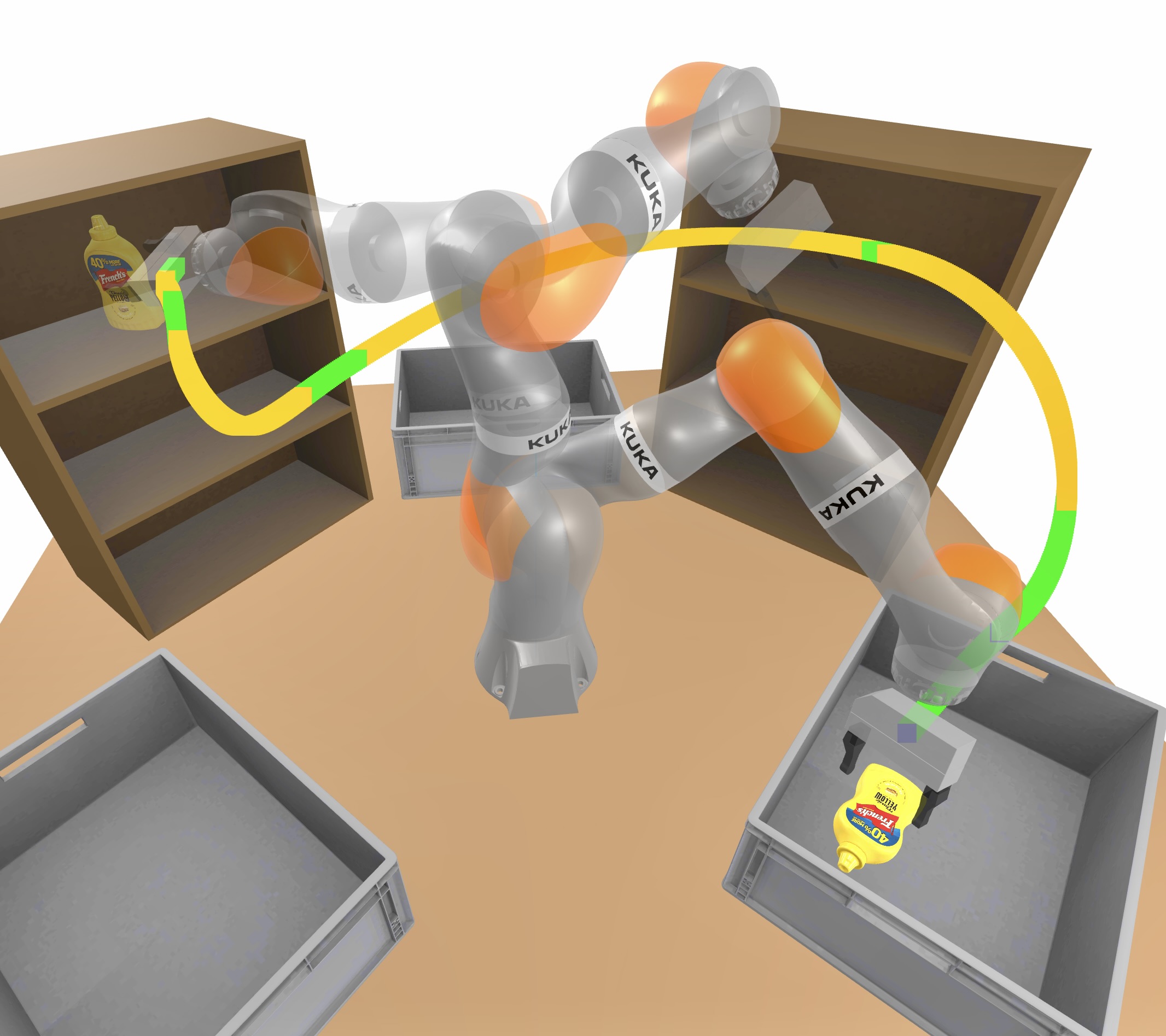}
            \caption{TOPP post-processes the trajectory to satisfy the acceleration limits, increasing its duration to 2.57s.}
            \label{sf-arm-spp-topp}
        \end{subfigure} &
        \begin{subfigure}[t]{0.31\textwidth}
            \centering 
            \includegraphics[width=\textwidth]{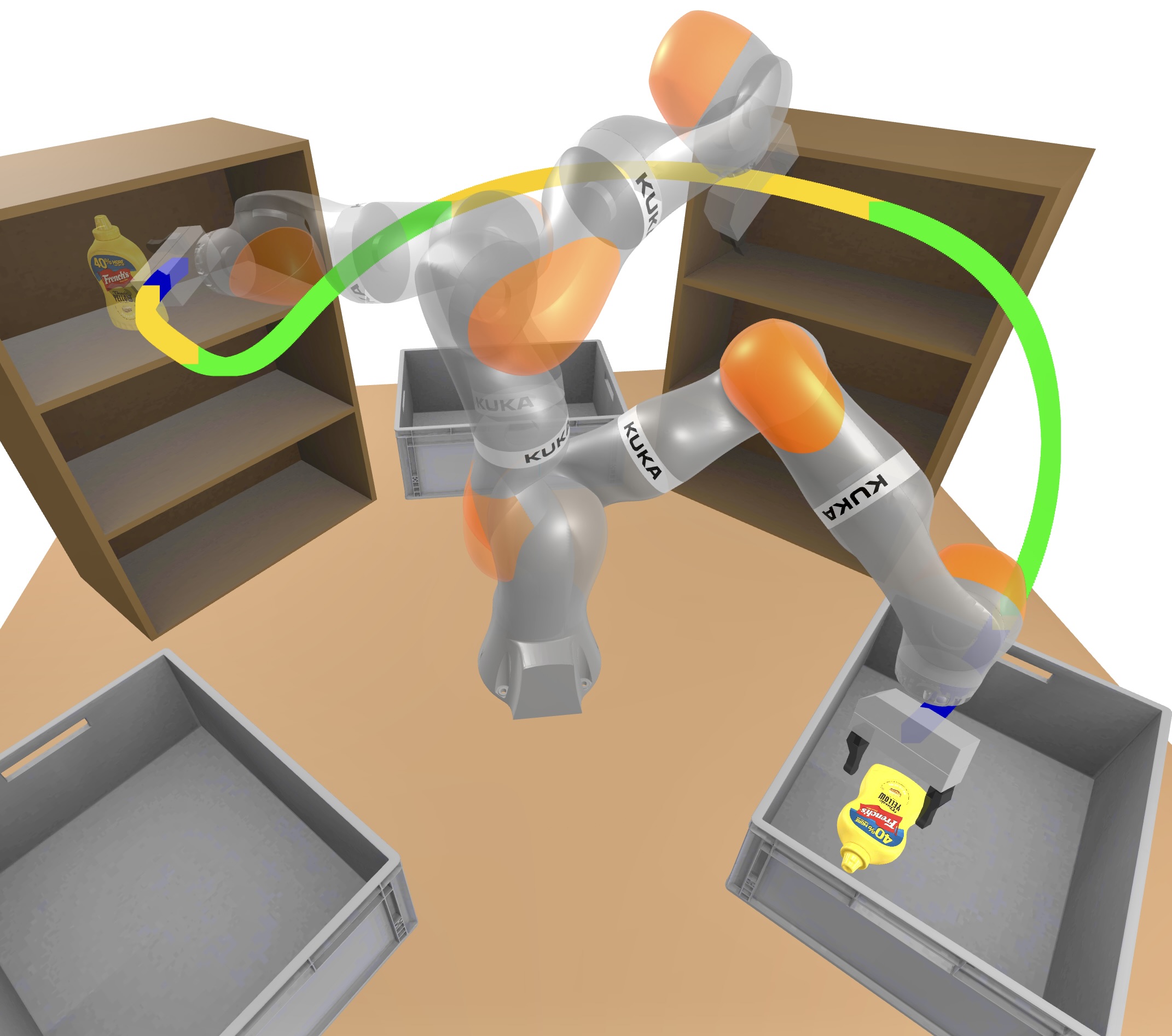}
            \caption{SWP in GCS enforces acceleration limits during search, yielding a faster 2.25s trajectory that operates at the limit longer.}
            \label{sf-arm-swp}
        \end{subfigure}
    \end{tabular}
    \caption{
    Visual comparison of SPP in GCS with TOPP (a,b) and SWP in GCS (c) formulations. 
    Both planners are tasked with finding an acceleration-limited trajectory from the top-left shelf to the right bin.
    The end-effector trajectory is red when the acceleration limit is violated for some joint, green when the acceleration is within 5\% of the limit, yellow when the velocity is within 5\% of the limit, and blue otherwise.
    }
    \label{f-robot-arms-collision-free}
\end{figure*}

To actually compute the optimal $(w,\tau)$, we solve multiple convex programs in parallel (one per candidate walk $w$) and select the best.
We take the first step of the selected walk, transitioning to $(v_{k+1}, x_{k+1})$, and proceed to next iteration.
If program \eqref{e-policy-spp-gcs} is infeasible, we backtrack to a previous vertex and try alternative candidates. 
The process terminates upon reaching the target $(t, \bar x_t)$ or exhausting a fixed iteration budget.
As such, this method lacks guarantees of completeness or optimality; still, we find it highly effective in practice.

To further improve the quality of solutions obtained via greedy search, we apply two post-processing steps.
First, we fix the walk and reoptimize the trajectory, yielding a solution that is optimal within the walk.
Next, we attempt to eliminate unnecessary cycles along the walk. 
If removing a cycle and re-optimizing the trajectory yields a feasible solution with a lower cost, we eliminate the cycle and continue.
Empirically, we observed this short-cutting technique to significantly improve the quality of solutions obtained via greedy search.

\section{Results}
\label{s-results}

We evaluate our shortest-walk formulation across several robotics systems to illustrate the framework's applicability and effectiveness.
These experimental demonstrations mirror the problem classes introduced in \Cref{s-many-problems-cast}.
All experiments are run on a laptop with an Apple M1 MAX chip with 16GB of RAM.
We use Mosek 10.2.1~\cite{mosek} to solve all convex programs in this section and SNOPT~\cite{gill2005snopt} to solve non-convex programs in section \Cref{ss-arm-collision-free-planning}, reporting parallelized 
solver time.

\subsection{Collision-free motion planning with acceleration limits}
\label{ss-arm-collision-free-planning}

We consider a robot arm in a warehouse setting, tasked with moving items between shelves and bins, illustrated in \Cref{f-robot-arms-collision-free}.
Given a pair of start and target conditions inside some shelf and some bin, the goal is to plan a minimum-time collision-free trajectory that satisfies acceleration limits due to the motors at the joints.
Following the discussion in \Cref{ss-collision-free-planning-theory}, we cast this problem as the SWP and the SPP in GCS, contrasting these formulations.

\begin{figure}[t!]
    \centering
    \includegraphics[width=0.95\columnwidth]{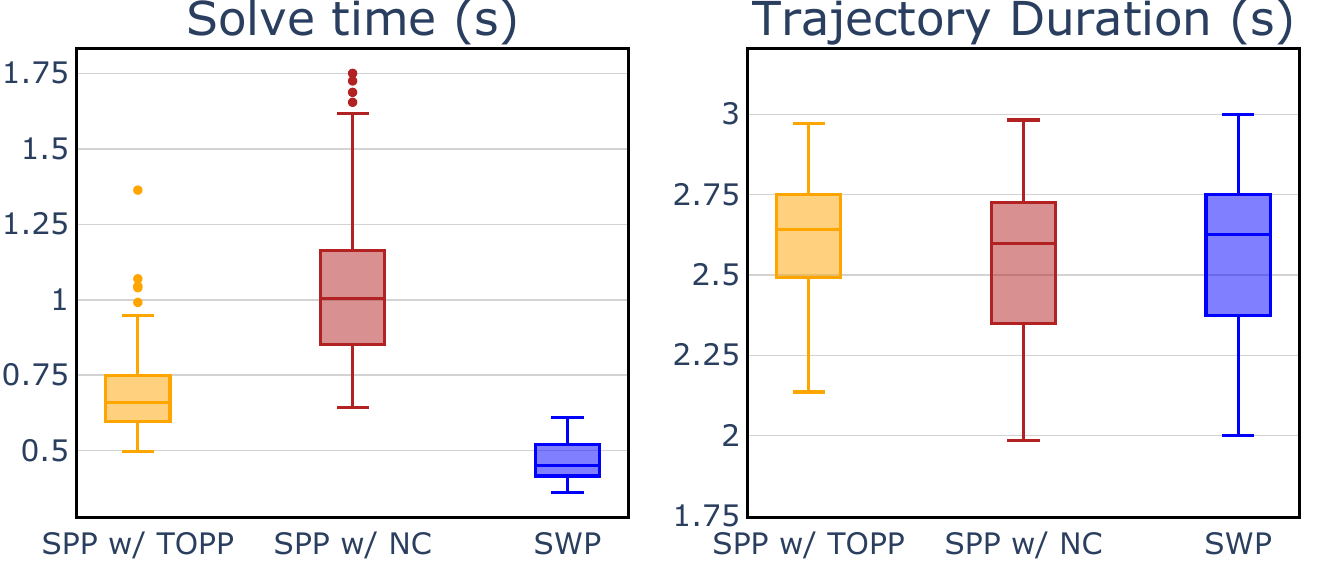}
    \caption{
    For collision-free planning scenario (\Cref{ss-arm-collision-free-planning}), we compare solve time and trajectory duration across 100 queries for shortest-paths with TOPP, shortest-paths with non-convex post-processing, and shortest-walk formulations. 
    SWP is 1.5 times faster than SPP with TOPP, 2.3 times faster than SPP with non-convex (left). 
    Trajectory durations are similar (right).
    }
    \label{f-numerics-robot-arm}
\end{figure}

First, we produce a convex decomposition of the collision-free configuration space using the IRIS-NP algorithm~\cite{petersen2023growing}.
In particular, we use IRIS clique seeding~\cite{werner2023approximating} to obtain regions that cover large volumes inside of the shelves and bins.
This one-time step takes 100 seconds, and the resulting connectivity graph has 16 vertices and 54 edges.
In what follows, we produce smooth trajectories through these collision-free regions, contrasting the shortest-path and shortest-walk formulations.

\begin{figure*}[t!]
    \centering
    \begin{tabular}{ccc}
        \begin{subfigure}[b]{0.31\textwidth}
            \centering
            \includegraphics[width=\textwidth]{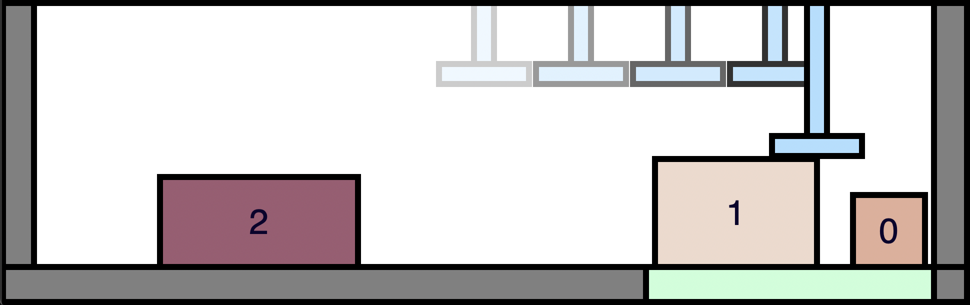}
            \caption*{1: grasp object 1.}
        \end{subfigure} &
        \begin{subfigure}[b]{0.31\textwidth}
            \centering
            \includegraphics[width=\textwidth]{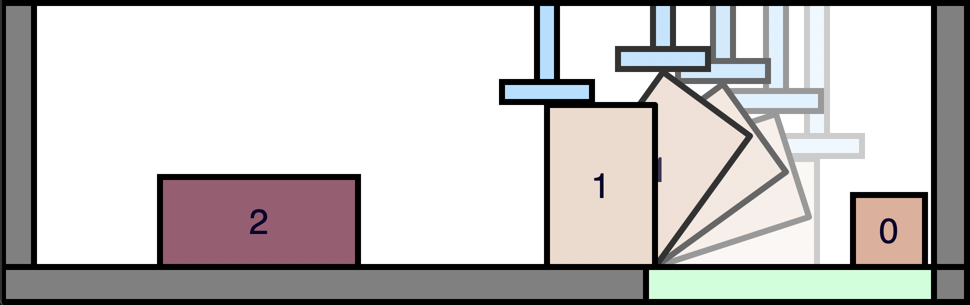}
            \caption*{2: flip object 1.}
        \end{subfigure} &
        \begin{subfigure}[b]{0.31\textwidth}
            \centering
            \includegraphics[width=\textwidth]{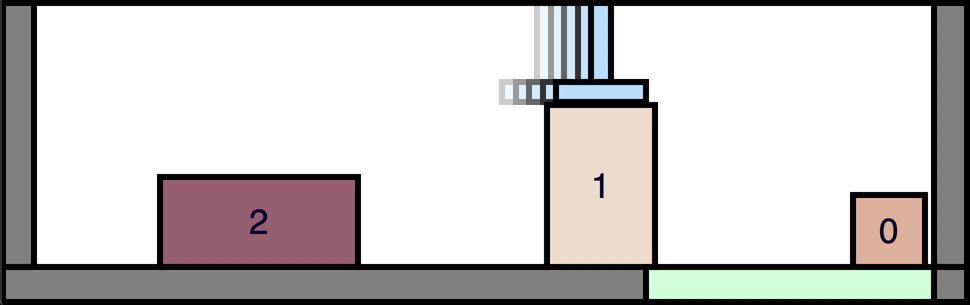}
            \caption*{3: regrasp object 1.}
        \end{subfigure} \\
        \begin{subfigure}[b]{0.31\textwidth}
            \centering
            \includegraphics[width=\textwidth]{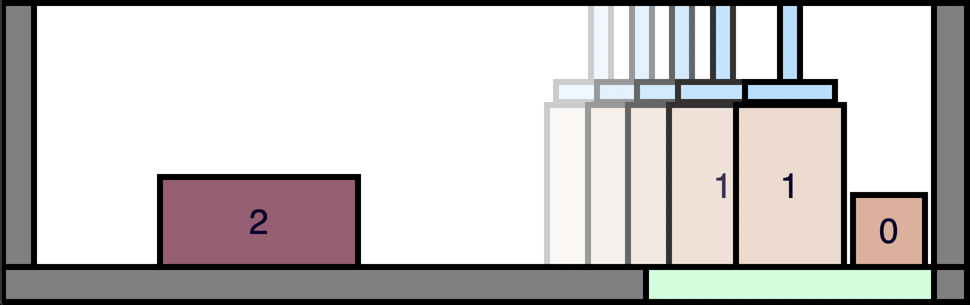}
            \caption*{4: move object 1 to target region.}
        \end{subfigure} &
        \begin{subfigure}[b]{0.31\textwidth}
            \centering
            \includegraphics[width=\textwidth]{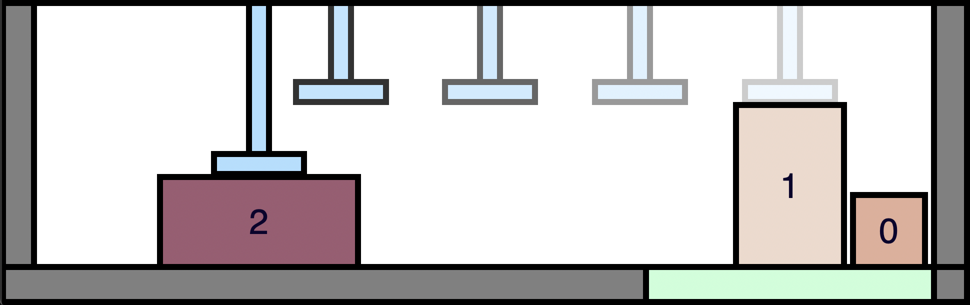}
            \caption*{5: grasp object 2.}
        \end{subfigure} &
        \begin{subfigure}[b]{0.31\textwidth}
            \centering
            \includegraphics[width=\textwidth]{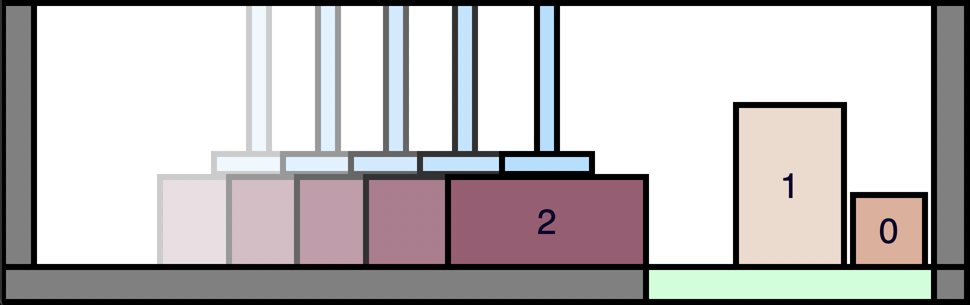}
            \caption*{6: move object 2.}
        \end{subfigure} \\
        \begin{subfigure}[b]{0.31\textwidth}
            \centering
            \includegraphics[width=\textwidth]{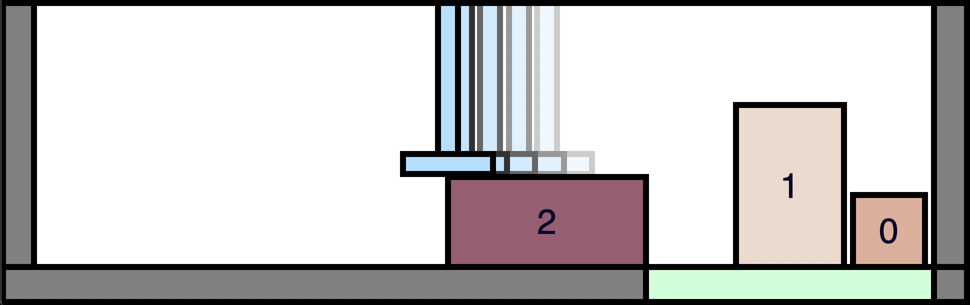}
            \caption*{7: regrasp object 2.}
        \end{subfigure} &
        \begin{subfigure}[b]{0.31\textwidth}
            \centering
            \includegraphics[width=\textwidth]{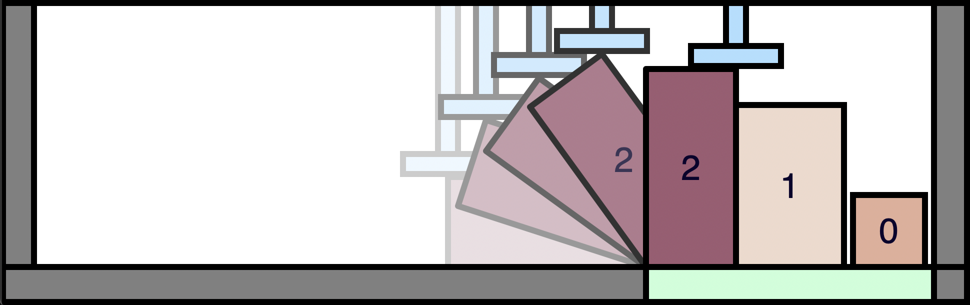}
            \caption*{8: flip object 2.}
        \end{subfigure} &
        \begin{subfigure}[b]{0.31\textwidth}
            \centering
            \includegraphics[width=\textwidth]{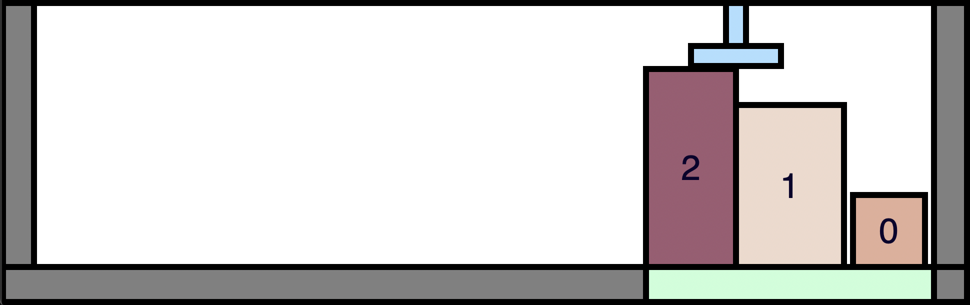}
            \caption*{9: task is complete.}
        \end{subfigure}
    \end{tabular}
    \caption{
    An 8-step plan where the top-down suction-cup arm is tasked with sorting three objects into the green target region.
    }
    \label{f-behavior-composition-results}
\end{figure*}

To cast the problem as the SPP in GCS, we follow the formulation from~\cite{marcucci2023motion}, jointly searching for the shape of the Bézier curve and its duration at every vertex, as depicted in \Cref{sf-bezier_spp_curves}.
The solution to the SPP in this GCS is a sequence of collision-free regions and a smooth velocity-limited trajectory through them.
Incorporating acceleration limits requires post-processing.
We consider two approaches: (1) TOPP~\cite{verscheure2009time}, which fixes the path and reoptimizes trajectory's timing to satisfy the limits; and (2) non-convex post-processing~\cite{von2024using}, which enforces the limits by solving a non-convex program over the selected sequence of collision-free regions.

For the SWP in GCS, we search for a cubic Bézier curve of fixed duration $\Delta t = 125\text{ms}$ at each vertex visit (see \Cref{sf-bezier_swp_curves}).
Fixing the duration allows us to explicitly enforce acceleration limits in a convex manner during incremental search. 
We compute the cost-to-go lower bounds over this GCS, which takes 45 seconds.
Per the discussion in \Cref{ss-synthesis-lower-bounds,ss-sdp}, we produce lower bounds that can be reused for multiple queries across the same GCS.

We conducted 100 tests where the arm moves virtual items between shelves and bins. 
The numerical results are shown in \Cref{f-numerics-robot-arm}, comparing the solve times and the durations of the produced trajectories.
As far as the solve-time goes, our SWP formulation (blue) takes on average 0.46s.
This is 1.5 times faster than the SPP with TOPP post-processing (orange) and 2.3 times faster than the SPP with non-convex post-processing (red).
As for the trajectory durations, the differences are marginal, within 1-2\% on average.

We illustrate the visual difference between SPP with TOPP post-processing and our SWP formulation in \Cref{f-robot-arms-collision-free}.
The SPP formulation first generates a velocity-limited trajectory of duration 2.21s, depicted \Cref{sf-arm-spp}.
This trajectory violates the acceleration limits, as indicated by the red segments in its end-effector trajectory.
TOPP post-processing reoptimizes the timing along the trajectory, producing a longer 2.57s trajectory that follows the same path but adheres to the acceleration limits, shown in \Cref{sf-arm-spp-topp}.
Observe that the red segments of the end-effector trajectory became green, reflecting that the acceleration is within 5\% of the limit.
In contrast, our SWP formulation directly enforces acceleration limits during search: it produces a faster 2.25s trajectory by operating at the acceleration limits for longer segments (green in \Cref{sf-arm-swp}).
Qualitatively, we also emphasize that TOPP cannot handle constraints on higher-order derivatives, including jerk, which our SWP formulation admits naturally.
This is particularly relevant for industrial robot arm applications.

Compared to SPP with non-convex post-processing, our SWP formulation produces trajectories of comparable duration but with much faster solve-time.
It also does not require installing an additional non-convex solver.

\subsection{Skill chaining}
\label{ss-behavior-composition-results}

We now consider a variation of the canonical pick-and-place problem, illustrated in \Cref{f-behavior-composition-results}.
Our robot is a suction-cup gripper with a fixed vertical orientation, described by a 1D horizontal position.
The environment contains three movable rectangular objects, described by their width, height, and the horizontal position.
The robot can pick and place objects using top grasps at their centers, or flip objects clockwise and counterclockwise by grasping their corners.
The objective is to sort all objects into a target region (a green horizontal interval).

\begin{figure*}[t!]
    \centering
    \begin{tabular}{cc}
        \begin{subfigure}[b]{0.48\textwidth}
            \centering 
            \includegraphics[width=\textwidth]{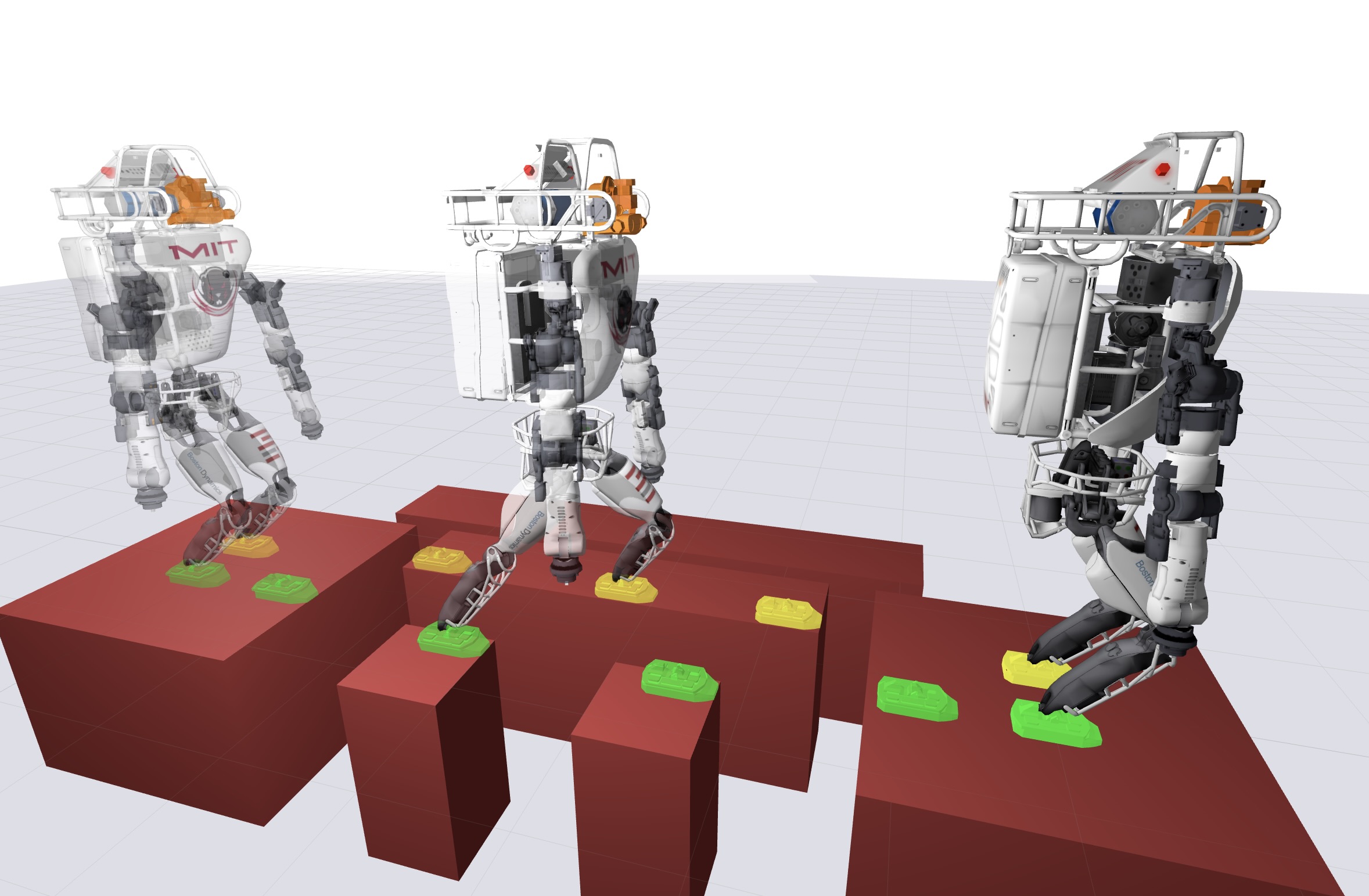}
            \caption{A footstep plan across a set of stepping stones.} \label{sf-atlas-walk-1}
        \end{subfigure} &
        \begin{subfigure}[b]{0.48\textwidth}
            \centering 
            \includegraphics[width=\textwidth]{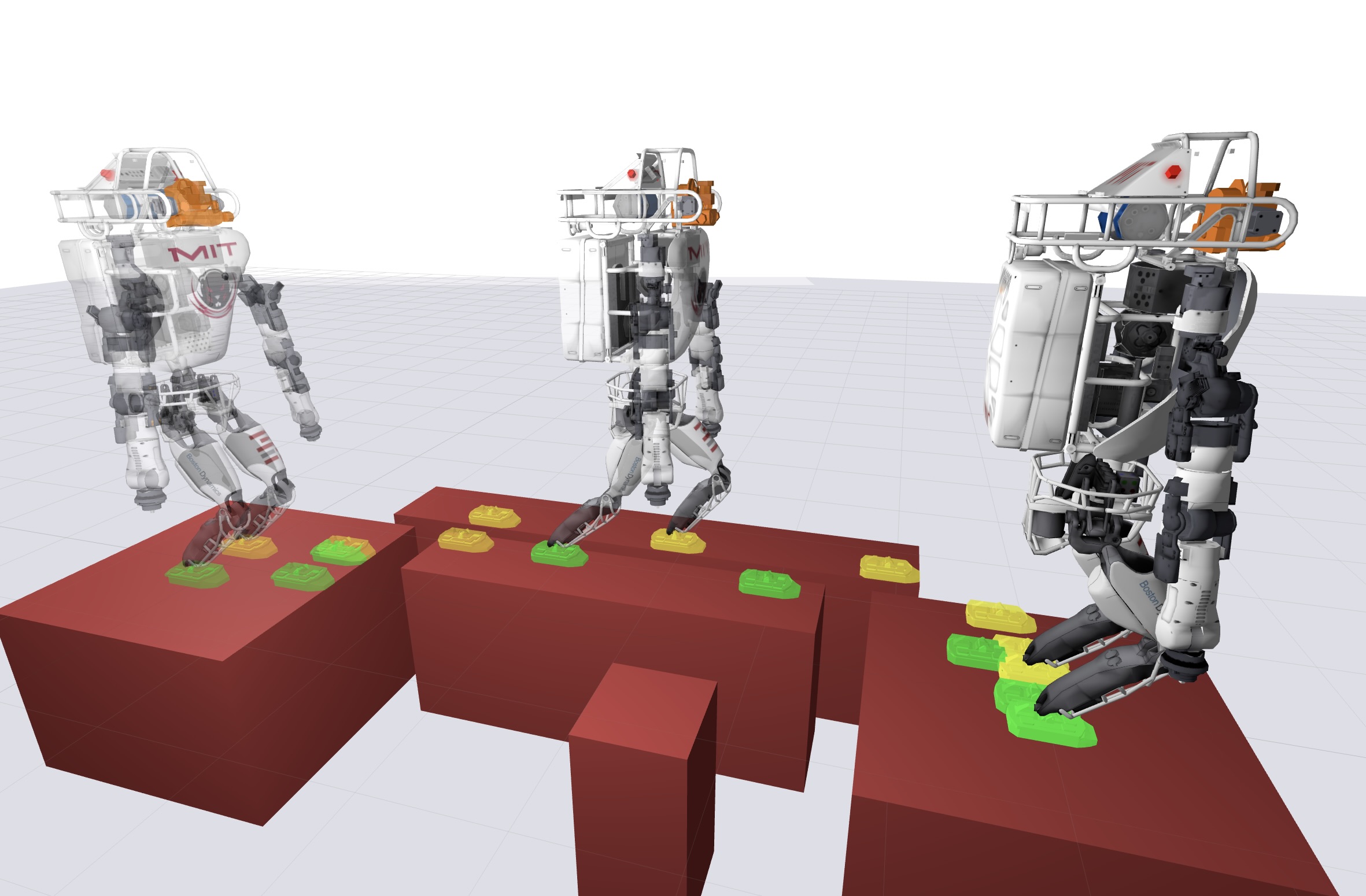}
            \caption{With one stone removed, the robot must take a longer detour.} \label{sf-atlas-walk-2}
        \end{subfigure}
    \end{tabular}
    \caption{
    Visualizations of the footstep plans across stepping stones for the Atlas bipedal robot.
    The SWP in GCS jointly optimizes for safe footstep placement, contact forces, and centroidal dynamics, while maintaining stability enforced by the ZMP condition.
    Both plans are produced in under 330ms.
    }
    \label{f-atlas-walking}
\end{figure*}

We cast this problem as a SWP in a GCS, following the skill chaining construction from \Cref{ss-behavior-composition-theory}.
We define the state space to be the horizontal positions of the arm and the objects, along with the objects' dimensions.
The robot's skills are defined as follows: 
(1) moving the arm from one position to another while the objects remain intact, 
(2) grasping an object at its center and placing it at a new collision-free location, and
(3) flipping an object clockwise or counterclockwise by grasping its corner, thereby swapping its width and height.
Skill execution cost is defined as 1 plus the arm's horizontal displacement, and the sets $\mathcal{X}_{\pi}$ are defined to capture the feasible transitions under each skill.
Naturally, these sets are not convex: this is due to the collision avoidance requirements and the combinatorial nature of selecting objects for manipulation.
To address this, we decompose the skills into convex sub-skills, resulting in 6 arm movement skills, 3 pick-and-place skills, and 6 object-flipping skills.
Our GCS contains a vertex for every sub-skill, a target vertex to represent any condition where the objects are inside the target region, and 6 source vertices to include any initial collision-free configuration of the objects.
The resulting GCS contains 22 vertices and 120 edges.
We maximize the average value of the quadratic cost-to-go lower bounds over the source vertices, which takes only 20 seconds.
The resulting structure efficiently supports shortest-walk queries for a variety of source conditions.

In \Cref{f-behavior-composition-results}, we demonstrate an example solution to the SWP in this GCS.
In this scenario, object 1 is already in the target region, but it needs to be moved and reoriented to also fit object 2.
This is a challenging puzzle, as all three objects barely fit within the target region.
For this reason, sampling-based planners would struggle to find any solution at all due to near-zero probability of sampling a feasible target configuration.
In contrast, our approach produces an optimal solution, avoiding unnecessary skill executions: for instance, object 0 remains in place, as it already satisfies the target conditions and doesn't need to be moved to fit the other objects.
With appropriate pre-building of the programs used during incremental search, the solve time for producing this and similar plans range from 0.5 to 1 second.

\subsection{Hybrid dynamics: footstep planning for a ZMP walker}
\label{ss-footstep-planning-results}

We now consider the problem of footstep planning for a bipedal robot navigating over stepping stones in a flat 2D $x$-$y$ plane. 
The robot, shown in \Cref{f-atlas-walking}, must reach the target by planning footsteps and contact forces through the stepping stones, ensuring stability and avoiding foot collisions.

We use the the well-known Zero-Moment Point (ZMP) formulation to model the robot's dynamics; see~\cite{kajita2003biped,vukobratovic2004zero} for classic reviews.
We constrain the ground reaction forces to lie inside the friction cone and also constrain the robot's Center of Pressure (CoP), which is also its ZMP, to remain within the support polygon formed by the feet.
This ensures that the robot can generate the ground reaction forces necessary to maintain dynamic stability.
We assume that the acceleration along the vertical axis is zero and that the robot does not rotate (i.e., zero angular acceleration).
This results in an affine relationship between the robot's CoP and CoM dynamics.
The dynamics become piecewise affine when we account for footstep planning, with three primary contact modes: both feet on the ground, only the left foot, or only the right foot. 
Foot placement on stepping stones introduces up to $O(N^2)$ potential contact modes ($N$ options for each foot), but many are infeasible due to constraints on the distance between the feet.
In practice, the modes scale as $O(DN)$, where $D$ is the average number of adjacent stones.
We assume that the feet do not rotate (this can be incorporated via McCormick envelopes~\cite{deits2014footstep} or SDP relaxations~\cite{graesdal2024towards}); additional constraints ensure collision avoidance between the feet.
The resulting PWA system jointly considers safe footstep placement, contact forces, centroidal dynamics, and ZMP-based stability.

Following the formulation in \Cref{ss-pwa-hybrid-dynamics-theory}, we cast discrete-time trajectory planning for this PWA system as the SWP in a GCS. 
A GCS vertex is added for each PWA mode, a target vertex is added to represent the desired location.
This results in a GCS with 24 vertices, 58 edges for the scenario in \Cref{sf-atlas-walk-1}, and 21 vertices, 50 edges for the scenario in \Cref{sf-atlas-walk-2}.
We compute quadratic cost-to-go lower bounds, maximizing their average value across all vertices, which takes 2.3 and 3.7 seconds respectively.
Guided by these bounds, finite-horizon greedy search generates an 10-step plan in \Cref{sf-atlas-walk-1} in 110ms and a 16-step plan in \Cref{sf-atlas-walk-2} in 330ms.
Given the speed of the greedy search, we note that the cost-to-go lower bounds can also be used to counteract disturbances and imperfect plan tracking, enabling receding-horizon replanning.

\section{Limitations}
\label{s-limitations}
Using GCS inherently requires the effort to construct it, which can be computationally intensive and may involve manual tuning, as was the case with generating collision-free regions with the IRIS algorithm in \Cref{ss-arm-collision-free-planning}, convex skill decomposition in \Cref{ss-behavior-composition-results}, and PWA decomposition of the dynamics in \Cref{ss-footstep-planning-results}.
Additionally, the cost-to-go synthesis step involves solving a potentially large SDP, taking several seconds to over a minute, which may be impractical for online planning. 
However, this one-time effort is justified if the cost-to-go lower bounds can be reused, such as in multi-query scenarios or as part of receding-horizon policy.
Finally, achieving the reported fast solve-times for the incremental search stage requires pre-building programs into binaries and solving them in parallel (we assumed we can solve up to 10 in parallel and reported simulated parallelized
solve-times).

\bibliographystyle{plainnat}
\bibliography{references}

\end{document}